\documentclass{scrartcl}
\usepackage[utf8]{inputenc}
\usepackage{amsmath}
\usepackage[english]{babel}
\usepackage[margin=1in]{geometry}
\usepackage{setspace}
\usepackage[authordate, backend=biber]{biblatex-chicago}
\usepackage{subfig}
\usepackage{graphicx}
\usepackage{makecell}
\usepackage{multicol}
\usepackage{booktabs}
\usepackage{caption}
\usepackage[utf8]{inputenc}
\RequirePackage[table]{xcolor}
\usepackage{subfiles}
\usepackage{multirow}

\definecolor{PANDarkGray}{RGB}{153,153,153}

\newcommand{\note}[1]{\footnote{\doublespacing #1}}
\title{Semantic Scaling: Bayesian Ideal Point Estimates with Large Language Models}
\author{Michael Burnham\footnote{Department of Political Science and Center for Social Data Analytics\\ Pennsylvania State University}\\mlb6496@psu.edu}
\date{October 2023 \\Word Count: 8291}
\begin{document}

\maketitle
\begin{abstract}
\noindent 
This paper introduces ``Semantic Scaling," a novel method for ideal point estimation from text. I leverage large language models to classify documents based on their expressed stances and extract survey-like data. I then use item response theory to scale subjects from these data. Semantic Scaling significantly improves on existing text-based scaling methods, and allows researchers to explicitly define the ideological dimensions they measure. This represents the first scaling approach that allows such flexibility outside of survey instruments and opens new avenues of inquiry for populations difficult to survey. Additionally, it works with documents of varying length, and produces valid estimates of both mass and elite ideology. I demonstrate that the method can differentiate between policy preferences and in-group/out-group affect. Among the public, Semantic Scaling out-preforms Tweetscores according to human judgement; in Congress, it recaptures the first dimension DW-NOMINATE while allowing for greater flexibility in resolving construct validity challenges.

\end{abstract}
\clearpage
\doublespacing

\section{Introduction}
\label{sec:intro}
The ideology of political actors is central to many models of political phenomena. Key to empirical tests of these models are reliable and valid measures for the beliefs of citizens, elites, or groups whose behaviors the theory seeks to explain. For this reason, scholars of legislatures \parencite{poole2009ideology}, judiciaries \parencite{martin2002dynamic}, elections \parencite{jessee2009spatial}, social media \parencite{barbera2015birds}, campaign finance \parencite{bonica2014mapping}, rebel groups \parencite{tokdemir2021rebel}, and more have all created unique measures of ideology.

Despite the fact that they seek to measure the same underlying concept, these models and methods generally treat estimating ideology as a unique task for the specific substantive application at hand. Often, this is a necessity because the measurement strategy relies on unique data, such as roll call votes, judicial decisions, surveys, expert opinions or manifestos, generated by the individuals or groups whose ideology the researchers seek to estimate. While these methods have enabled many tests of important theories, they also suffer from two notable weaknesses: First, they produce scores that are difficult to compare across political systems and institutions. This is because estimates are often derived from data sources peculiar to an institution. As a result, a sub-literature on how to ``bridge" between ideal spaces has developed \parencite[e.g.][]{jessee2016can, epstein2007judicial}. 

Second, most approaches place individuals on undefined ``left-right" scales. This may be sufficient for many applications, but advances in the study of ideology have created a need to measure the concept in different ways. While concerns about the dimensionality of ideology have long been raised \parencite[e.g.][]{lauderdale2014scaling, poole2009ideology}, scholars have recently turned their attention to nonpolicy-related ideological considerations, most notably affect. Scholars now routinely differentiate between affective (i.e. in-group love and out-group hate) and ideological (i.e. policy preference) polarization \parencite[e.g.][]{iyengar2012affect, iyengar2019origins, druckman2019we, costa2021ideology}. Work by \textcite{mason2015disrespectfully, mason2018uncivil} shows groups can display high levels of affective polarization while having similar policy preferences. Ideal point estimates that do not differentiate between affect and policy thus may not be appropriate for many research questions. 

Current methods of measuring affect include self reports and surveys \parencite[e.g.][]{iyengar2012affect, druckman2019we}, implicit association tests \parencite{iyengar2015fear}, and behavioral measures such as bias in dating and labor markets \parencite{huber2017political, gift2015does}. Most notably, the feeling thermometer ``has become the primary vehicle for measuring affect toward a wide range of groups in the electorate" \parencite{iyengar2019origins}. These approaches are all expensive to implement and potentially limited in the populations they can reach. While they have had great success measuring affect among citizens, it is difficult to translate these them to political elites and groups. As a result, the affective aspects of political beliefs remain difficult to study. 

I propose a new method, useful across contexts and actors, to measure ideology. Nearly all citizens, politicians, and political groups publicly express their beliefs. In the process, they produce text which can be used to measure their ideology. Text is often generated at a high frequency – providing scholars a regular stream of new data. Further, text provides a more explicit signal about affect, policy preferences, or other elements of ideology than do voting behavior, campaign contributions, or implicit association tests.

However, while current approaches that estimate ideology from text are fast and convenient, they have inherent limitations that prevent them from fully capitalizing on the aforementioned advantages of text. First, they often rely on word counts or word vectors for scaling -- methods that struggle to make detailed semantic inferences. Second, many are restricted to specific populations or document types. Finally, they scale across all word or embedding features within the text to derive ideal points rather than allowing researchers to explicitly define the features. These limitations prevent the creation of more well specified scaling instruments such as those that explicitly measure affect or policy preferences, or topic specific scales such as a measurement on attitudes towards gun control or approval of party leadership.

The approach described in this paper, called Semantic Scaling, relies on large language models to classify documents based on their semantics or meaning. From these classifications, I extract survey-like data from observed subjects. With these data, I use Bayesian Markov Chain Monte Carlo techniques to estimate subjects' ideological positions. This approach can be used to measure the ideologies of citizens, elites, and groups; it allows researchers to define the type of ideology to be estimated (e.g., policy-based or affective) and it provides robust estimates on a variety of document types and length.

I validate the method using two examples from American politics: the policy preferences of Twitter users and the policy and affective ideological positions of members of Congress. First, I demonstrate that Semantic Scaling recaptures the results of Tweetscores \parencite{barbera2015birds}, the leading approach to estimating user ideology from Twitter. Notably, I find that when Tweetscores and Semantic Scaling disagree, human labelers tend to agree with Semantic Scaling. Second, I show that Semantic Scaling produces policy-based ideology scores that match DW-NOMINATE \parencite{poole2009ideology}. Further, I demonstrate that because Semantic Scaling allows researchers to explicitly define their ideological dimensions it can credibly measure legislators’ in-group/out-group affect. This represents the first scaling method to credibly measure affect from passively observed data and provides opportunities for scholars to measure affect for populations -- like elites and groups -- for whom survey-based measures of the concept are elusive.

\section{Text Scaling Methods}
With the many potential advantages of ideal point estimation from text, it is no wonder that Wordfish and Wordscores significantly impacted the political science literature \parencite{slapin2008scaling, laver2003extracting, lauderdale2016measuring}. In the right context, these approaches, as well as more recently developed alternatives \parencite[e.g.][]{lauderdale2016measuring, temporao2018ideological} provide fast and robust estimates of ideology with widely accessible software implementations. A more recent approach proposed by \textcite{rheault2020word} advances these word-based scaling methods by using word vectors, rather than word counts, and party labels to create ``party embeddings" for scaling legislators.

However, research validating these methods demonstrates that robust estimates are limited to narrow contexts. \textcite{temporao2018ideological} find that Wordfish does not work well for social media data, and \textcite{grimmer2013text} reach similar conclusions on legislator press releases. \textcite{bruinsma2019validating} similarly find that Wordscores fails to capture estimates for European parties consistent with expert judgement, and Party Embeddings from \textcite{rheault2020word} are confined to legislators by design. In all cases, text based scaling methods get relatively poor separation between parties compared to gold-standard methods such as DW-NOMINATE \parencite{lauderdale2016measuring, rheault2020word}.

Both the premise and challenge of current text scaling methods is that they attempt to create a task-specific language model via word vectors or word counts that computers can use to infer semantics. This has two primary drawbacks. First, it necessitates a large volume of data to construct a sufficient model. In the case of word counts, it also necessitates the questionable assumption that context can be safely ignored. Second, the process of placing all aspects of language use into a vector space and then scaling from the resulting model is perhaps less principled than would be ideal. Scaling is a dimensionality reduction task and scaling from a holistic model of language means it is often unclear exactly what linguistic features are influencing point estimation. This implies that scaling methods may leverage regionalisms or other linguistic features that correlate with ideology, but are not ideological.\note{Consider a toy example with the statement ``What do y'all think about Biden?''. The phrase conveys no ideological information. However, if language is modeled using word counts and co-occurrences ``y'all" might be incorrectly associated with negative views on Biden because it is used more frequently in the American South. Because voters in the American South tend to have more negative views on Joe Biden, southern regionalisms will more-frequently co-occur with negative statements about Biden and be assumed to communicate negative sentiments towards him.} Careful text pre-processing and document selection may somewhat alleviate this challenge, but does not eliminate it. These pre-processing steps also introduce multiple arbitrary decision points that can significantly alter results \parencite{denny2018text}.

As a result, current text-based scaling methods do not fully capitalize on the promise of text as a data source for ideal point estimation. They provide great utility in the right circumstances because they are easy to implement and can readily incorporate new data. However, they are either particular to a population or document type or produce dimensions that cannot be readily defined.

A potential alternative approach to using word counts or word vectors is to organize documents by topic, label them for the political preferences they express based on their semantics or meaning, and then use a traditional item response theory (IRT) model to infer ideology. This would circumvent many of the challenges inherent to word counts or vector based methods. Rather than scaling from word counts or the undefined dimensions of a word vector, it would estimate ideology from expressed beliefs -- similar to survey responses. 

However, this introduces a new problem of scope. Labeling the documents must be automated for the approach to be practical and computers cannot natively understand document semantics to produce such labels. Supervised classifiers may work on a small scale, but labeling documents for each item we wish to scale along (e.g. presidential approval, out-group attitudes, gun control, etc.) might entail compiling many training data sets and training many classifiers.

Recently developed language models solve this problem of scope. They are increasingly adept at inferring the semantics of documents without additional training \parencite[e.g.][]{laurer2022annotating, burnham2023stance,  gilardi2023chatgpt}. This process of using a generalized language model to label data without training it is known as ``zero-shot" classification. The implication is that researchers can use a single model to obtain high quality labels on an arbitrary number of items of interest with no manual labeling or model training. This largely obviates the need to compile large volumes of data, long documents, or estimate the distribution of word counts to infer the position a document expresses. 

Thus, new advances in large language models enable a shift in the paradigm of how we can estimate ideology from text. Rather than deconstructing language use into word count distributions or modeling the meaning of individual words with vectors to estimate ideology, we can instead label documents for the political positions they express and apply a more conventional IRT model to the document labels to produce ideal point estiamtes. This allows researchers to explicitly define which features are contributing to ideal point estimates, separate scales of interest such as affect and policy preferences, estimate the preferences of citizens, elites, and groups, and obtain estimates with relatively small amounts of data.

\section{Approach}
A useful touch point for conceptualizing the process of Semantic Scaling is an open text survey. Consider a survey that asks for short text answers to a variety of political questions (e.g., ``In politics, as of today, do you consider yourself a Republican, a Democrat or an independent?", ``In general, do you feel that the laws covering the sale of firearms should be made more strict, less strict or kept as they are now?). A researcher might then label responses as conservative or not and use a two parameter logistic item response theory model to measure ideology.

The general population, political elites, and organizations all routinely provide information akin to these survey responses via social media, press releases, opinion articles, and more. Members of Congress send newsletters to their constituents expressing their views on the issues before them. Citizens share their opinions on social media. Interest Groups write opinion pieces and press releases to stake their positions publicly. All of these expressions of position are text ripe for analysis. By labeling these documents for expressed beliefs, they can be used as data in a similar item response theory model to produce a measure of ideology. 

However, text corpora often present at least two additional challenges that complicate the modeling process. The first is the aforementioned problem of scope. Unlike in a survey, subjects in an open source data context can generate multiple text samples related to a single item. This adds an additional dimension to the number of text samples that must be labeled. Formally, the number of text samples is $n \times k \times \mu_k$, where $n$ is the number of research subjects, $k$ is the number of items or topics used for scaling, and $\mu_k$ is the average number of text samples users generate per item. This number of documents and items will quickly become so numerous that manual labeling and supervised classifiers are too resource intensive for most research applications.

The second challenge is that this process produces count data -- and in some instances highly dispersed count data. Conventional IRT models, factor analysis, or other scaling methods are not designed for such data. Transformation of the data from count to either binary or continuous [0,1] data would constitute a significant loss of information: what items people choose to engage with more frequently is potentially ideologically significant. In the next two sections I present solutions to each of these challenges.

\subsection{Labeling Data}
The initial challenge is to identify which stances each of the documents in our corpus express. While a comprehensive overview of stance identification in text is beyond the scope of this paper, I outline a general process explained by \textcite{burnham2023stance} by way of example. This frames stance detection in terms of what is known as ``entailment classification." In entailment classification, text samples are classified in ``statement'' and ``hypothesis'' pairs. The statements are the corpus of documents we wish to label, and the hypotheses are statements created by the researcher that represent a particular stance. The classifier then determines if the hypothesis is true, given the content of a document. Consider, for example, an op-ed headline and hypothesis pair such as the following:

\hspace{8pt} \textbf{Statement:} ``I'm Joining the Republican Party. Here's Why."

\hspace{8pt} \textbf{Hypothesis:} ``The author of this text supports Republicans."

\noindent The standard for textual entailment is that a statement entails a hypothesis if a human reading the statement would conclude that the hypothesis is true \parencite{aldayel2021stance}. Thus, in this example the correct classification is true and we would count this as one observation of the author supporting Republicans. To label our corpus we simply repeat this process for all documents and hypotheses that we are interested in.

%Contemporary large language models are particularly adept at such classification tasks, even in a zero-shot context \parencite{yin2019benchmarking, laurer2022annotating, burnham2023stance}. This enables researchers to quickly label documents based on the semantics rather than relying on sentiment or dictionary based approaches that often return labels uncorrelated with the stances expressed \parencite{aldayel2021stance, bestvater2022sentiment, burnham2023stance}. 

The objective of classifying documents in this fashion is to generate an $n \times k$ matrix that contains counts of the number of times an author $n$ expressed opinion $k$. If we wish to scale attitudes towards Republicans, for example, we might have a column in our matrix that counts the number of documents expressing support for Republicans, and another column that counts the number of documents expressing opposition to Republicans. It should be noted that longer documents can contain many ideologically significant statements. Accordingly, in most cases it will be appropriate to break up longer documents into sentences or paragraphs for classification and thus a ``document" may only be a single sentence. A typical pipeline for labeling documents might consist of the following:
\begin{enumerate}
    \item Tag documents according to which scaling items they are related to. This may be done by keywords or topic models. For example, I might assume all documents containing the word ``republican" are related to the dimensions measuring attitudes towards Republicans.
    \item Create a set of hypotheses that represent the opinions you want to classify for each scale (e.g. ``The author of this text supports Republicans", ``The author of this text opposes Republicans").
    \item Match documents to relevant hypotheses via keywords or topic labels and classify documents using a large language model for entailment of hypotheses. 
\end{enumerate}

To accomplish the classification step, researchers have a number of language models to choose from. While GPT-4 is currently the most capable stance classification model available \parencite{burnham2023stance}, I want to avoid using large proprietary models that are not reproducible and are expensive to use at a large scale. Accordingly, I use an open source transformer model trained specifically for entailment classification, also known as natural language inference \parencite{yin2019benchmarking}. These models are open source, significantly more accessible and reproducible than GPT-4, and achieve classification performance similar to supervised classifiers \parencite{burnham2023stance}. 

While my approach represents an acceptable compromise between accessibility and current state-of-the-art, the landscape of language models is rapidly evolving. More sophisticated models will become increasingly accessible and this method will improve as classification performance improves. I emphasize that the objective is to simply obtain a matrix with the count of documents generated that express a set of stances. Researchers should consult recent literature to determine which approaches and models are most appropriate for their data.

\subsection{Assumptions and Statistical Model}
I assume the documents authors generate are more likely to entail hypotheses they agree with than hypotheses they do not. This assumption is consistent with spatial models of ideology that underlie scaling methods based on roll call votes \parencite{poole2009ideology}, campaign contributions \parencite{bonica2014mapping}, and network analysis \parencite{barbera2015birds}.

However, modeling ideology from document counts poses a particular challenge. To obtain accurate estimates we cannot assume the count of conservative (liberal) documents is solely a function of ideology. An ideologically moderate person with a high rate of document generation may create more conservative documents than a conservative with low document generation. This precludes using established count IRT models, such as the Conway-Maxwell Poisson IRT model \parencite{beisemann2022flexible}, designed for testing or experimental conditions where subjects are assumed to have similar levels of engagement.

To solve this, we can assume document generation is a Bernoulli process in which each document person $i$ generates is a conservative (liberal) document about item $j$ with probability $p_{ij}$. The count of person $i$'s conservative (liberal) documents about item $j$, $x_{ij}$, is thus binomially distributed:
\begin{equation}
    x_{ij} \sim B(y_i,p_{ij})
    \label{eq_binom}
\end{equation}
Where $y_i$ is the total count of documents generated. \note{An alternative parameterization of the model assumes $x_{ij} \sim B(y_{ij},p_{ij})$. That is, $x_{ij}$ is binomially distributed around the total count for item $j$ rather than simply the total document count. This assumes observations have an equal probability to generate content about item $j$ and thus the topics people engage with is not ideologically relevant.}

Other political science applications of IRT to measure ideology model the probability that an action (e.g. vote) is conservative (liberal)\parencite[e.g.][]{martin2002dynamic}. We can then think of $p_{ij}$ in the above sampling distribution similarly. The probability that a document is about item $j$ and is conservative (liberal), $p_{ij}$, can be expressed as as a two parameter IRT model in slope-intercept form as:
\begin{equation}
    p_{ij} = logit^{-1}(\delta_j + \alpha_j\theta_i)
    \label{eq_p}
\end{equation}
Where $j$ is the item or document topic, $\delta_j$ is the intercept for item $j$, $\alpha_j$ is the discrimination parameter, and $\theta_i$ is person $i$'s ideology. By substituting equation \ref{eq_p} in to equation \ref{eq_binom} the sampling distribution can be defined as a logit parameterization of the binomial distribution:
\begin{equation}
    x_{ij} \sim B(y_{i}, logit^{-1}(\delta_j + \alpha_j\theta_i))
\end{equation}
By substituting $logit^{-1}(\delta_j + \alpha_j\theta_i)$ for $p_i$ in the cumulative distribution function for the binomial distribution we can derive the likelihood function for $n$ observations and $k$ items as:
    \begin{equation}
    \mathcal{L}(\delta,\alpha,\theta|x,y) = \prod_{i=1}^n \prod_{j=1}^k \binom{x_{ij}}{y_{i}} logit^{-1}(\delta_j + \alpha_j\theta_i)^{y_{i}}(1-logit^{-1}(\delta_j + \alpha_j\theta_i))^{x_{ij}-y_{i}}
    \label{eq:likelihood}
\end{equation}

This model is unidentified for maximum likelihood estimation -- because the $ \alpha_j\theta_i$ term is multiplying two unknown variables, the variables can take on many values while still creating a valid solution. In lieu of maximum likelihood, we can use Bayesian methods to identify the model with a few assumptions. By constraining theta to $N(0,1)$ we can identify the scale and location of the parameters  \parencite{bafumi2005practical, gelman2007data}. However, the model still suffers from reflection invariance, meaning the estimated ideological scale can be flipped without a change to the prediction. As a result, it is not clear which ideological direction is represented by positive or negative values. This can be resolved by leveraging the prior information we have about observations to set initial values for $\theta$ in the expected directions \parencite{bafumi2005practical}. For example, if party labels are available we could assign Democrats an initial $\theta$ value of -1 and Republicans a value of 1. Where party labels are not available, we can use information known about the topics discussed in our documents. In the applications below, I assign people that make more conservative statements than liberal statements a prior value of 1 and all other observations are given a prior of -1.

With these constraints, we can use maximum a posteriori (MAP), or Bayesian MCMC to fit the model and obtain ideal point estimates. The joint posterior of the distribution can be specified as:

\begin{equation}
    p(\delta, \alpha, \theta|x, y) \propto \mathcal{L}(\delta,\alpha,\theta|x,y) \Pi[p(\alpha_j) p(\delta_j) p(\theta_i)]
\end{equation}
where $p(\alpha_j)$, $p(\delta_j)$, and $p(\theta_i)$ are the assumed prior probability density functions of the parameters (e.g. normal, Cauchy). If researchers have additional information on the structure of ideology in their sample this can readily be incorporated. For example, researchers may wish to incorporate random effects if party labels are known. In the applications for this paper, I use MCMC with No-U-Turn sampling to obtain estimates with standard errors and use standard normal priors on all parameters. This identifies the model with weakly informed priors and ensures that results shown do not represent overfitting to strong priors.

\subsection{Items and Data Structure}
The nature of document counts introduces a final complication. Conventional IRT models assume item labels are mutually exclusive -- a vote, ruling, survey answer, etc. can be either liberal or conservative but not both. Under this assumption, item responses can be given a binary coding of conservative (liberal) or not. 

This assumption does not hold for document counts. If each item can be answered multiple times it can be answered in both liberal and conservative directions.\note{For example, a person may tweet that they are ``pro-life" and also that they oppose a ban on abortion at six weeks.} Further, documents about political topics need not express any particular political preference, such as a news headline stating factual events. This creates a potential problem in estimating the ideology of moderates if items are only represented once with a count of liberal or conservative documents. To resolve this, each item or topic on the scale should be represented twice in the data. Once as a count of conservative documents and a second time as a count of liberal documents.

\section{Application 1: Political Twitter}
Social media is a particularly fruitful medium for Semantic Scaling. It is a common space for both political elites and the general public, and many platforms produce large volumes of political text in the form of posts, titles, captions, etc. Social media text is also often short in length, which can pose a challenge for word counts methods because it can result in sparse word count matrices.

\subsection{Data and Methods}
I use Twitter as an initial application because it is a well researched medium for political discussion and is thus ideal for testing new methods. Tweetscores -- a network-based method of scaling -- has been validated against party registration and NOMINATE scores \parencite{barbera2015birds} and thus provides a useful benchmark. For each user I obtain ideal point estimates for both affect and policy preferences and test if Semantic Scaling is able to recapture the ideological distribution measured by Tweetscores.

I gathered a sample of 32,000 politically active Twitter users and collected all of their tweets generated between September 1, 2020 and February 28, 2021. I then estimated ideal points for each user with the Tweetscores package in R \parencite{barbera2015birds}. From the initial sample I culled observations for which I could not reliably estimate ideology with Tweetscores ($\hat{R} > 1.05$). Slight attrition also occurred during the data collection period due to users being banned or deleting their accounts. At the end of collection the sample consisted of 21,628 Twitter users. I then randomly sampled 10,000 users that had at least 15 posts related to affect and 15 posts related to policy preferences.

To define the affective scale, I follow the literature from \textcite{amira2021group}, \textcite{iyengar2019origins}, \textcite{druckman2019we} and others. Specifically, I conceptualize affect as in-group love and out-group hate. Affective documents are those that express opinions about either political elites (politicians or media personalities) or political groups (liberals and conservatives or Republicans and Democrats). I distinguish between general references to political groups and references to media and political elites based on the findings by \textcite{druckman2019we} that people differentiate between attitudes towards elites and ordinary citizens. I further differentiate between elites and the two presidential candidates because the presidential candidates are disproportionately discussed compared to other elites and occupy a distinct position as party leaders in an election year.
 
To define aspects of the policy preference scale, I used a topic model to create a list of twenty common political topics discussed during this period. I eliminated topics I determined did not capture clear policy divides (e.g. ``the economy") and selected the five most commonly discussed topics as determined by document counts. A list of the items associated with each scale is found in Table \ref{tab:summary_stats}. As mentioned in section 3.3, items are represented three times in my data: once for liberal statements and once for conservative statements. This makes for a total of 22 items across the two scales. Items are not considered mutually exclusive and thus any document may count towards both affective and policy scales.  

The topics each tweet is associated with was determined via simple keyword matching. A list of keywords for each item is in Appendix C. Table \ref{tab:summary_stats} provides a summary of the distribution of each item. Affective topics were much more commonly discussed than policy, with the exception of COVID-19 response. We also see that the count data exhibits extreme overdispersion.

\begin{table}[!htbp]
  \centering
  \begin{tabular}{@{\extracolsep{5pt}} llrrr}
    \hline \hline
    Scale & Item & Median & Mean & Variance \\
    \hline
    \multirow{6}{*}{Affect}  & Trump & $132$ & $499$ & $1,451,823$ \\
     & Biden & $30$ & $122$ & $91,822$ \\
     & Conservatives & $62$ & $237$ & $359,892$ \\
     & Liberals & $23$ & $104$ & $81,497$ \\
     & Conservative Elites & $20$ & $79$ & $46,634$ \\
     & Liberal Elites & $31$ & $108$ & $71,375$ \\\hline
    \multirow{5}{*}{Policy} & Election Fraud & $8$ & $48$ & $25,535$ \\
     & COVID-19 & $55$ & $172$ & $157,048$ \\
     & Guns & $5$ & $16$ & $1,303$ \\
     & Abortion & $2$ & $7$ & $421$ \\
     & Race and Policing & $3$ & $11$ & $605$ \\
    \hline
  \end{tabular}
  \caption{Tweet count distribution across items}
  \label{tab:summary_stats}
\end{table}

I created a basic set of entailment hypotheses that represent conservative, liberal, and neutral stances for each item and used these with an entailment classifier to label documents. A complete set of these hypotheses is in Appendix C. Hypotheses were matched to tweets based on the key words they contained. I then used a DeBERTa model trained for zero-shot entailment classification to label document-hypothesis pairs \parencite{laurer2022annotating, he2020deberta}. The result is an $n \times k$ matrix that contains the counts of conservative and liberal tweets made by each user along each item.
%Items on the affect dimension follow a simple template:
%\begin{itemize}
%    \item The author of this text supports (person/group)
%    \item The author of this text opposes (person/group)
%    \item The author of this text expresses no opinion about (person/group)
%\end{itemize}
%Where the (person/group) field is populated with either the name of a politician or political group. Hypotheses for policy preferences generally follow a similar template:
%\begin{itemize}
%    \item The author of this text believes (thing) is good
%    \item The author of this text believes (thing) is bad
%    \item The author of this text believes (thing) is neutral
%\end{itemize}
%Where (thing) is replaced by some policy or policy related item (e.g. vaccines, voter IDs, Roe v. Wade). A few alternative hypotheses were created where the above template was not appropriate (e.g. ``The author of this text believes systemic racism exists/does not exist"). 

Finally,  I used Bayesian Markov Chain Monte Carlo with No-U-turn Sampling and standard normal priors on the parameters to estimate the model outlined in equation \ref{eq:likelihood}. To solve reflective invariance I use the procedure outlined above and assign a prior of 1 to $\theta_i$ for individuals with more conservative tweets and a prior of -1 to $\theta_i$ for individuals with more liberal tweets. I then estimated ideology along affective and policy scales independently, and then again combining affect and policy items into a single scale. The model ran for 40,000 iterations after burn-in and no parameter had an $\hat{R}$ higher than 1.01. Trace plots and additional convergence diagnostics are in Appendix D.

\subsection{Results}
In Figure \ref{fig:twit_mat}, I report the correlations and their associated standard errors between Semantic Scales, as well as plot the distributions and scatter plots. All scales recapture the bimodal distribution of the data and both the joint policy and affect scale is highly correlated with Tweetscores ($\rho = 0.9$, $se = 0.004$). The affective scale is notably more informative to the combined scale than policy, and the affective scores correlate more highly ($\rho = 0.9$, $se = 0.004$) with Tweetscores than the policy scores do ($\rho = 0.82$, $se = 0.006$). The local minimum between the modes on the policy scale is notably higher than any of the other measurements, indicating lower levels of polarization on policy. While this could potentially be alleviated by adding more policy related items to the scale, it is also consistent with \textcite{mason2015disrespectfully}, which find that policy preferences can cut across political identities.

\begin{figure}
    \centering
    \includegraphics[width=.9\textwidth]{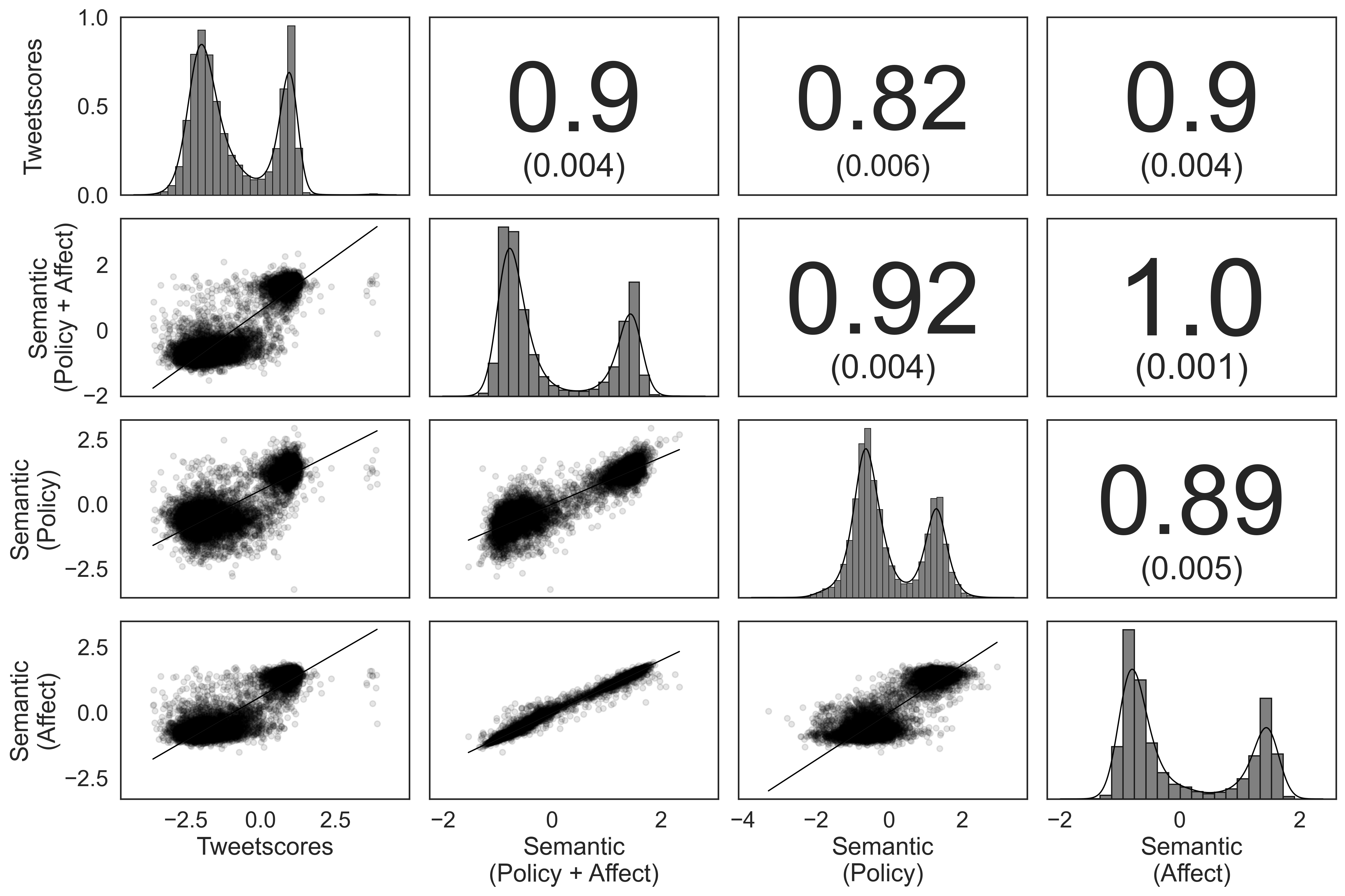}
    \caption{Semantic Scaling recaptures the bimodal distribution and is highly correlated with Tweetscores.}
    \label{fig:twit_mat}
\end{figure}

\begin{figure}
    \centering
    \includegraphics[width = \textwidth]{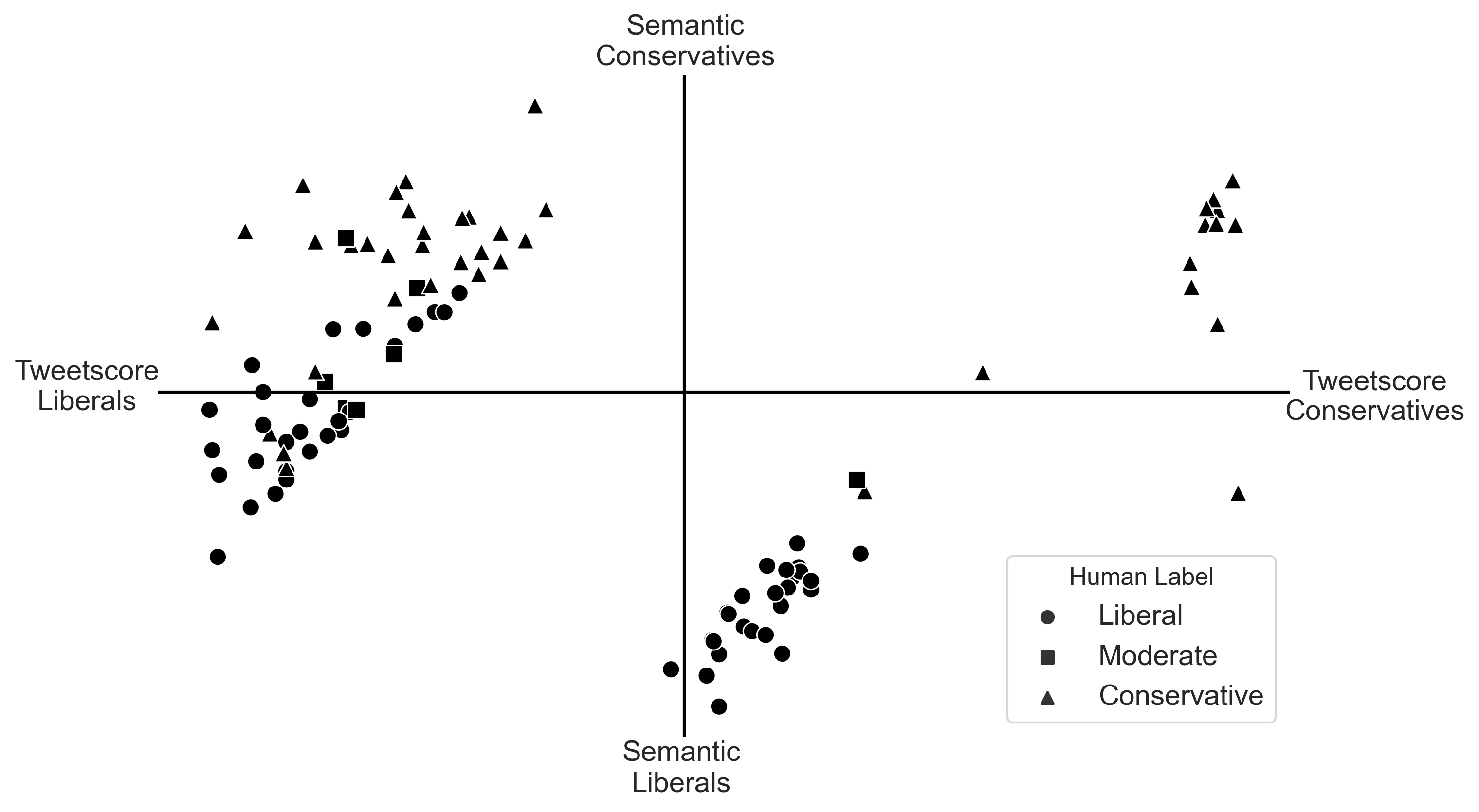}
    \caption{Scatter plot of the 100 observations with the highest residuals when regressing Tweetscores on Semantic Scaling. Axes are drawn at the local minimum between the modes of the distributions to roughly divide liberals and conservatives. Human labels derived by reading a random sample of tweets from each user largely agree with Semantic Scaling over Tweetscores.}
    \label{fig:residuals}
\end{figure}

When the two methods disagree, which is right? To answer this question I examined observations where Tweetscores and Semantic Scaling differ the most. To select these points I regressed Tweetscores on the joint Semantic Scale with a simple bi-variate linear model and selected the 100 observations with the highest residuals. Figure \ref{fig:residuals} shows a scatter plot of these observations. The axes are drawn at the local minimum between the modes of both the Tweetscores (x-axis) and Semantic (y-axis) distributions to roughly divide liberal and conservative observations. Observations in quadrants two and four indicate directional disagreements in ideology between the methods while the observations in quadrants one and three represent disagreements of magnitude.

For each of these observations, I pulled a random sample of 25 political tweets and had a research assistant read the tweets and determine if they thought the author was liberal, conservative, or moderate. The results indicate that on points of disagreement, Semantic Scaling provides superior estimates of ideology according to human judgment. Semantic Scaling is correlated with human labels at $\rho = 0.75$ ($se = 0.07$) while Tweetscores is correlated at $\rho = 0.18$ ($se = 0.10$). In quadrants one and three, we see that Tweetscores, Semantic Scaling, and human labels largely agree. In quadrants two and four, where Tweetscores and Semantic Scaling have directional differences, human labels largely agree with Semantic Scaling.

\section{Application 2: The 117th Congress}

Estimating the ideologies of members of Congress provides a second point of validation. DW-NOMINATE scores, which are commonly used measures of Congressional ideology, rely on members' votes to generate ideal point estimates. As \textcite{cox2007legislative} have argued, the set of votes that members take on the floor is highly curated by partisan agenda setting. This, coupled with other aspects of legislative politics--such as strategic abstentions--suggest that the roll call record is not an ideal data source for unbiased estimates of legislative ideology \parencite{carrubba2008legislative, rosas2015no, ainsley2020roll}.

Estimating Congressional ideologies offers a second point of validation via DW-NOMINATE scores. While \textcite{poole2009ideology} argue roll call votes represent true preferences, they also acknowledge that their analysis of House committees and roll call votes is consistent with \textcite{cox2007legislative}'s model of party behavior in Congress. This model suggests parties control votes and minimize defection primarily through controlling which bills make it to the floor for a vote. Thus, while roll call votes provide a wealth of ideologically informative data points, the information contained in the data is necessarily bounded by the agenda setting priorities of the parties. This implies that minority factions that lack agenda setting power may not be well represented by DW-NOMINATE scores while pivot points within Congress that have significant agenda setting power will have robust ideological estimates. Congress thus provides an interesting point of variance to see if Semantic Scaling can capture aspects of ideology not well represented by roll call votes.

\subsection{Data and Methods}
The data I use consists of all newsletters sent by the 117th Congress collected by the DC Inbox project \parencite{cormack2017dcinbox}, as well as all tweets sent by members and their campaign Twitter accounts. To demonstrate that Semantic Scaling does not require Twitter data, Appendix 1 compares semantic scaling with Wordfish \parencite{slapin2008scaling} and finds that Semantic Scaling produces consistent results regardless of the data source. Here, I present results that use both newsletter and Twitter data because the additional data provides slightly more robust estimates and allows me to include more members of Congress. I subset the data to only include members of Congress that had at least five documents related to the affective and policy scales, for a minimum of ten documents. Documents are defined as either an entire tweet, or a single sentence from a newsletter. This resulted in a total of 433 members of Congress and 425,028 documents. To pre-process the documents I removed non-alphanumeric characters except emoji and divided newsletters into discrete sentences using a simple rule based sentence splitter \parencite{sadvilkar-neumann-2020-pysbd}. 

To classify the data for stance I used a process identical to the first application. Items related to affect consist of statements about party leadership, presidential candidates, or Republicans (or conservatives) and Democrats (or liberals) as a group. To identify policy items I first scraped the list of most viewed bills for the 117th Congress from Congress.gov. I then counted the number of times each bill was mentioned in the documents and derived a list of the ten most frequently discussed bills. I then classified sentences mentioning these bills as either supporting or opposing them. I limited the list of bills to ten because I found that counts of documents mentioning bills beyond the most popular were very sparse, with even the most popular bill only mentioned in ~3\% of documents. The bills among the most discussed primarily concerned infrastructure spending and COVID-19 relief stimulus, and the Respect for Marriage Act. In addition to these bills, I added general discussion on abortion, border control, and gun control to the policy preference scale as these were frequently discussed topics not well represented by the most popular bills.

I then compiled a list of hypotheses for entailment classification and matched hypotheses to documents using keywords (e.g. all documents containing the word ``abortion" were classified for stances expressed about abortion.) A complete set of items and their associated hypotheses are in Appendix C. The Bayesian model ran for 100,000 iterations and no parameter had an $\hat{R}$ higher than 1.01. Complete convergence diagnostics are in Appendix D.

\subsection{Results}
\begin{figure}
    \centering
    \includegraphics[width=.9\textwidth]{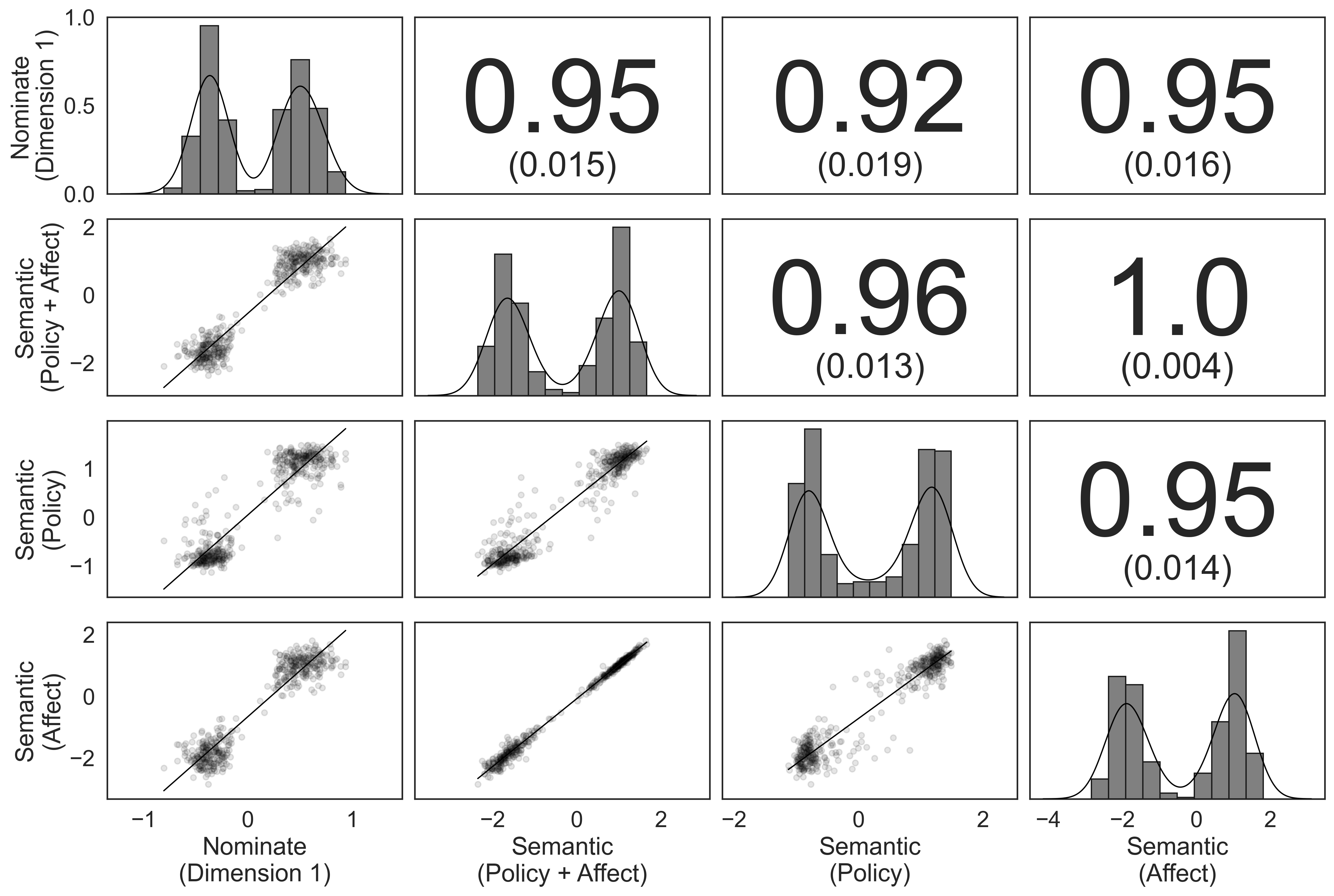}
    \caption{The correlations and standard errors between Semantic Scaling and DW-NOMINATE demonstrate that Semantic Scaling is able to recapture the distribution of DW-NOMINATE.}
    \label{fig:cong_corr}
\end{figure}

Ideal point distributions and correlations with DW-NOMINATE are shown in Figure \ref{fig:cong_corr}. Semantic Scaling is highly correlated with DW-NOMINATE ($\rho = 0.95$, $se = 0.015$) and recaptures the bimodal nature of the distribution. Some additional demonstrations of face validity may be informative here. To do so, I examine two groups in Congress: Senate moderates with agenda setting power and ``The Squad" and ``MAGA Squad" sub-factions among House Democrats and Republicans.

\subsubsection{Senate Moderates}
\begin{figure}
    \centering
    \includegraphics[width=\textwidth]{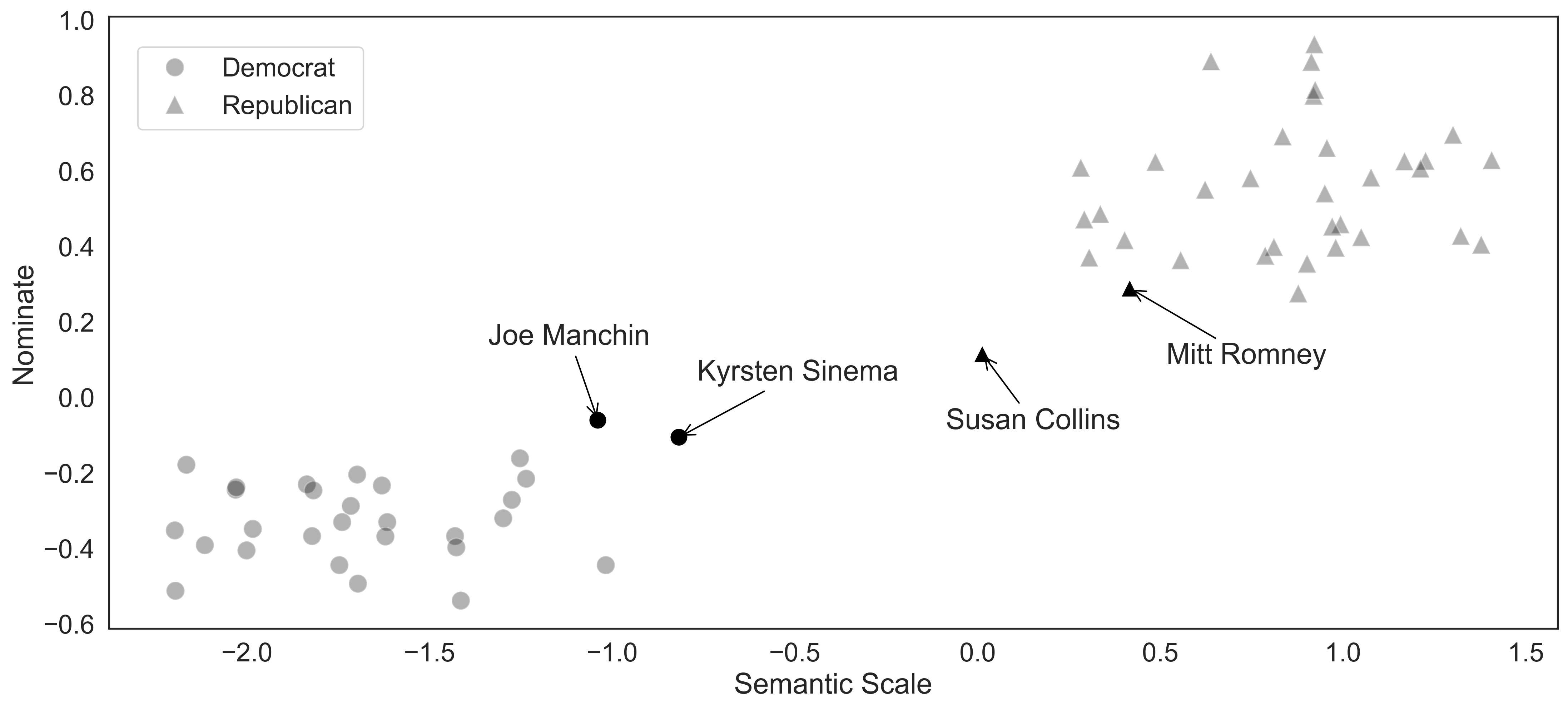}
    \caption{Both DW-NOMINATE and Semantic Scaling predict Mitt Romney, Susan Collins, Kyrsten Sinema, and Joe Manchin to be among the most moderate members of the Senate.}
    \label{fig:Senate_scatter}
\end{figure}

 Senate moderates offer a unique point of validation because of the slim majority parties maintained during the 117th Congress. In accordance with \textcite{krehbiel2010pivotal}'s pivotal politics model, moderates have substantial power in agenda setting when slim majorities are maintained. Roll-call-based scaling should thus provide robust estimates of moderate's ideologies as votes are more likely to reflect the preferences of pivot points. Figure \ref{fig:Senate_scatter} plots DW-NOMINATE scores against Semantic Scaling with Joe Manchin, Susan Collins, Kyrsten Sinema, and Mitt Romney highlighted -- four senators with a reputation for moderation and agenda control.\note{Lisa Murkowsi lacked sufficient data to be included in the analysis.} DW-NOMINATE identifies these senators among the most moderate in the chamber. Semantic Scaling produces nearly identical results.

\subsubsection{House Sub-factions}
In contrast to the Senate moderates, ``The Squad" (widely considered to consist of Alexandria Ocasio-Cortez, Ilhan Omar, Ayanna Pressley, Rashida Tlaib, Jamaal Bowman, and Cori Bush in the 117th Congress) poses a significant scaling challenge for DW-NOMINATE. This small coalition of Democrats has a reputation for being highly progressive. However, \textcite{lewis2022} notes that, as of 2019, this coalition of liberal House members are estimated as moderates by DW-NOMINATE. While the authors note they expect The Squad members to drift to the left over time, this drift has not manifest as of 2023. 

% If The Squad are moderates according to DW-NOMINATE, one might expect House Republicans that similarly break with their leadership to also be measured as moderate. However, Matt Gaetz, Marjorie Taylor Greene, Lauren Boebert, and Paul Gosar (hence forth the ``MAGA Squad") similarly broke with party leadership and are estimated as some of the most conservative members of the House according to DW-NOMINATE.

This is the so-called ``ends against the middle" problem occurs where extreme ends of the party break with the majority because they are not liberal or conservative enough. \textcite{duck2022ends} propose a statistical model that deftly resolves this issue in some contexts. My goal here is not to propose another solution to the ends against the middle problem, but rather to point out that the problem is one of construct validity. Conceptualized as a measurement of how extreme policy preferences are, DW-NOMINATE fails relative to The Squad while \textcite{duck2022ends} succeeds. Conceptualized as a measurement of which of the two parties members prefer to vote with, DW-NOMINATE succeeds while \textcite{duck2022ends} fails.

\begin{figure}
    \centering
    \includegraphics[width=\textwidth]{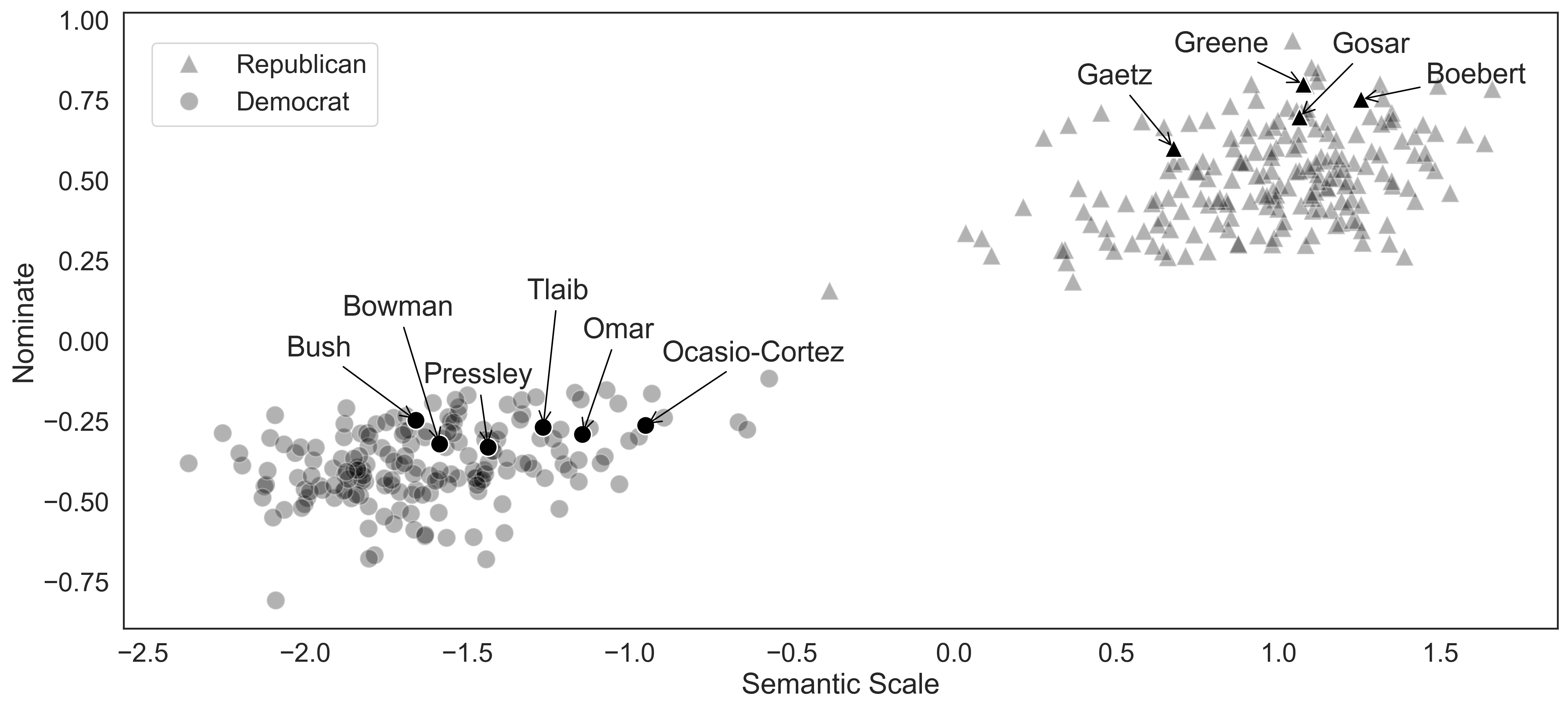}
    \caption{Both DW-NOMINATE and the Semantic Scale with all items show similar results for House subfactions. The MAGA Squad is clearly conservative while the Squad is perhaps more moderate than would be expected.}
    \label{fig:house_scatter}
\end{figure}

There are two key points here: First, construct validity is contingent upon the researcher's measurement goals. And second: Scaling is a dimensionality reduction technique and results may not align with research goals when items are indiscriminately added to the model. As shown in Figure \ref{fig:house_scatter}, Semantic Scaling actually produces similar estimates to DW-NOMINATE for House sub-factions when all items are included in the model. Members of The Squad are generally further right than most would anticipate. As demonstrated below, this is due to the fact that members of the squad more frequently criticize Democratic leadership than their colleagues. The implication is that a single ideal point estimate like DW-NOMINATE or some other metric may not be appropriate across all research questions. Rather, it may be better to specify a scale that ensures minimal information loss across the dimensions of interest.

Semantic Scaling offers more leeway in aligning measurement and research objectives by allowing researchers to more readily define the contents of their ideological scale. Consider a research question in which ingroup-outgroup animosity is the treatment variable. Rather than relying on existing metrics like DW-NOMINATE that may be misaligned with our goals, we can instead use Semantic Scaling to construct something akin to a ``feeling thermometer" for the parties.

\begin{figure}%
    \centering
    \subfloat{{\includegraphics[width=5cm]{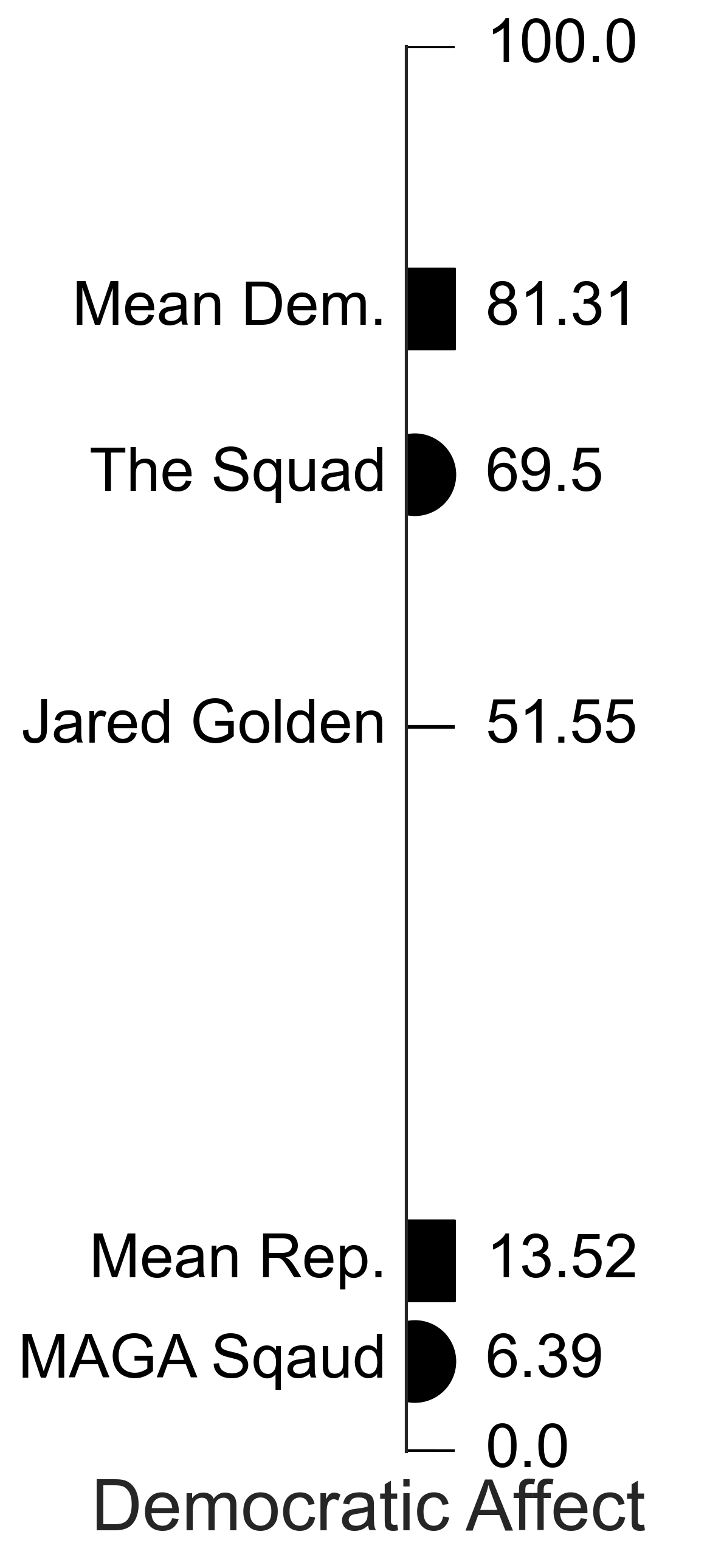} }}%
    \qquad
    \subfloat{{\includegraphics[width=5cm]{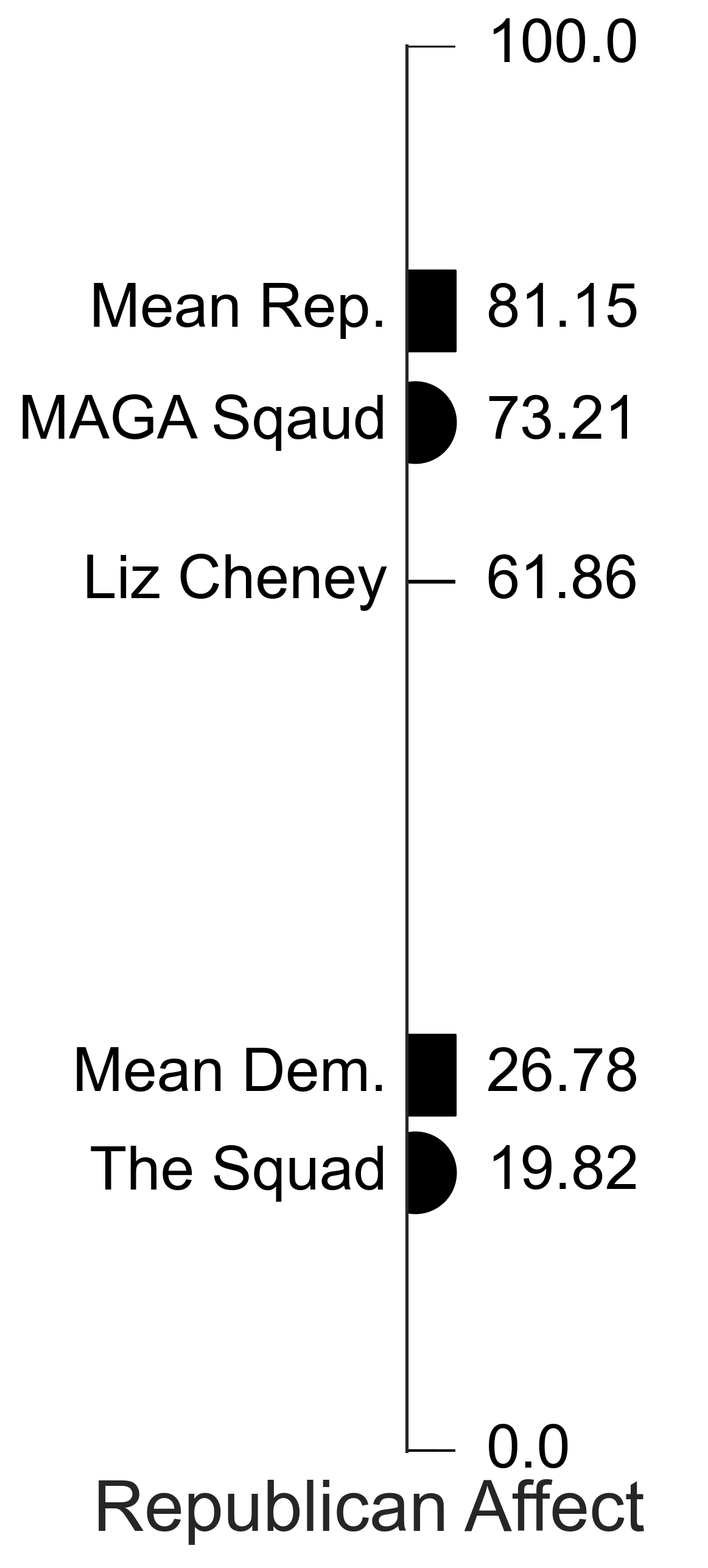} }}%
    \caption{Party affect similar to feeling thermometers constructed by scaling across items related to the parties and party leadership, then applying a linear transformation on the results so that the lowest observed value is equal to zero and the maximum observed value is equal to 100.}%
    \label{fig:thermom}%
\end{figure}

In Figure \ref{fig:thermom}, I use the items related to affect towards both parties in order to create thermometers for the Republican and Democratic parties. Zero represents the lowest oberserved value and 100 represents the highest observed value, similar to a feeling thermometer. Consistent with expectations, I find that the Squad and the MAGA Squad are more extreme than their average party member in terms of out-group animosity. However, we also find that they have less positive affect for their own party than the average party member. This is consistent with their propensity to criticize party leadership. As additional points of reference, I include two outspoken critics of their own parties: Jared Golden for the Democratic party (current leader of the Blue Dog Democrat coalition) and Liz Cheney for the Republican Party (perhaps the most vocal dissident of her party with regards to the Trump impeachment trials and the January 6th insurrection).

% \begin{figure}
%   \centering
%   \subfloat{\includegraphics[width=\textwidth]{Figures/dem_love_bw.png} \label{fig:demlove}} \\
%   \subfloat{\includegraphics[width=\textwidth]{Figures/rep_love_bw.png} \label{fig:replove}}
%   \caption{One-dimensional plots showing attitudes towards Republicans and Democrats on a love-hate scale.Both the Squad and the MAGA Squad show higher than average negative affect towards the out-party but lower than average positive affect toward the in-party} \label{fig:lovehate}
% \end{figure}

\section{Discussion}
This paper presented Semantic Scaling, a text-based method of ideal point estimation from stance classification. The approach uses large language models to classify stance across scales defined by the researcher and then an item response theory model to infer ideology. The approach is flexible, works well with short texts and low document counts, and can capture aspects of ideology that may not be captured well through other mediums such as social networks or roll call votes. Perhaps most significantly, it allows researchers to explicitly define the dimensions along which they wish to scale. This paper demonstrates that Semantic Scaling can differentiate the between affective and policy aspects of ideology, but future research may apply it to arbitrary aspects of ideology.

The approach is not without challenges, however. Unlike methods such as Tweetscores or DW-NOMINATE, Semantic Scaling does not scale across pre-defined items or data. Rather, Semantic Scaling is more akin to scaling from a survey in that researchers must define the number and content of items with each application. This provides great flexibility, but makes validation across applications more challenging. Further, text-based methods of ideology estimation may struggle in instances where actors are incentivized to obfuscate their preferences.

As language models continue to advance in both capabilities and efficiency, Semantic Scaling will be increasingly accurate and accessible. Future research should endeavor to build software and models that streamline this process. The results presented in this paper use models trained for general language understanding. While these models are sufficient for a wide variety of applications, domain adaptation to political speech can provide significant performance benefits \parencite{kawintiranon2022polibertweet, burnham2023stance}. Tuning classifiers to labeled stance detection data sets could thus enable better zero-shot classification performance for political scientists and further improve the reliability and flexibility of Semantic Scaling. 

Substantively, research in affective polarization is ripe for applications in Semantic Scaling due to its ability to differentiate between group attitudes and policy preferences. This allows researchers to move beyond simple sentiment dictionaries and adopt a more robust measurement approach in a wider variety of communication contexts than other scaling methods would allow.

%\clearpage
% \noindent \large{\textbf{Funding}}

% \normalsize
% \noindent None.
% \\ \\
% \noindent \large{\textbf{Acknowledgements}}

% \normalsize
% \noindent I would like to thank Michael Nelson, Burt Monroe, Kevin Munger, and Suzie Linn for their helpful comments and guidance during this research. I am grateful to Justin Grimmer, Santiago Olivella, as well as other participants at the PolMeth 2022 annual conference for their constructive feedback on an earlier draft of this article.
% \\ \\
% \noindent \large{\textbf{Data Availability Statement}}

% \normalsize
% \noindent Data and replication materials for this article are forthcoming.

% %\noindent \large{\textbf{Supplementary Material}}

% % \normalsize
% % \noindent Tutorials for implementing the methods presented in this article are available at \url{https://github.com/MLBurnham/stance_detection_tutorials}

% \noindent \large{\textbf{Conflicts of Interest}}

% \normalsize
% \noindent None.

\clearpage

%\bibliographystyle{chicago-authordate}
%\bibliography{refs}
\printbibliography
%\clearpage
%\appendix
%\subfile{Sections/appendix}

\end{document}

\tableofcontents
\clearpage
\doublespacing
\section{Wordfish Comparison}
\begin{figure}[h]
    \centering
    \includegraphics[width = 0.9\textwidth]{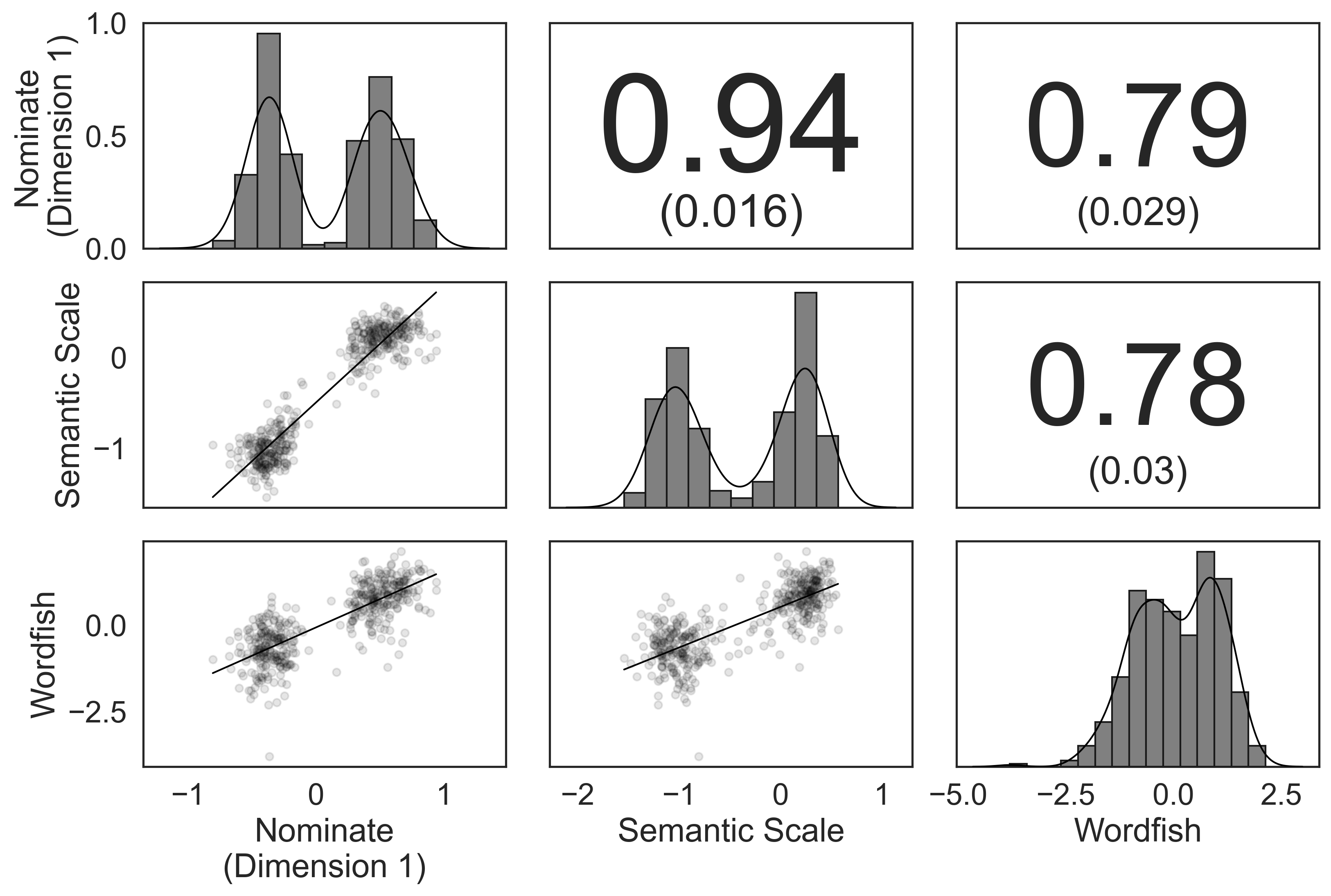}
    \caption{Scaling results using newsletter data only. Wordfish does not adequately capture the bimodal distribution of congress.}
    \label{fig:fish_letters}
\end{figure}

\begin{figure}[h]
    \centering
    \includegraphics[width = 0.9\textwidth]{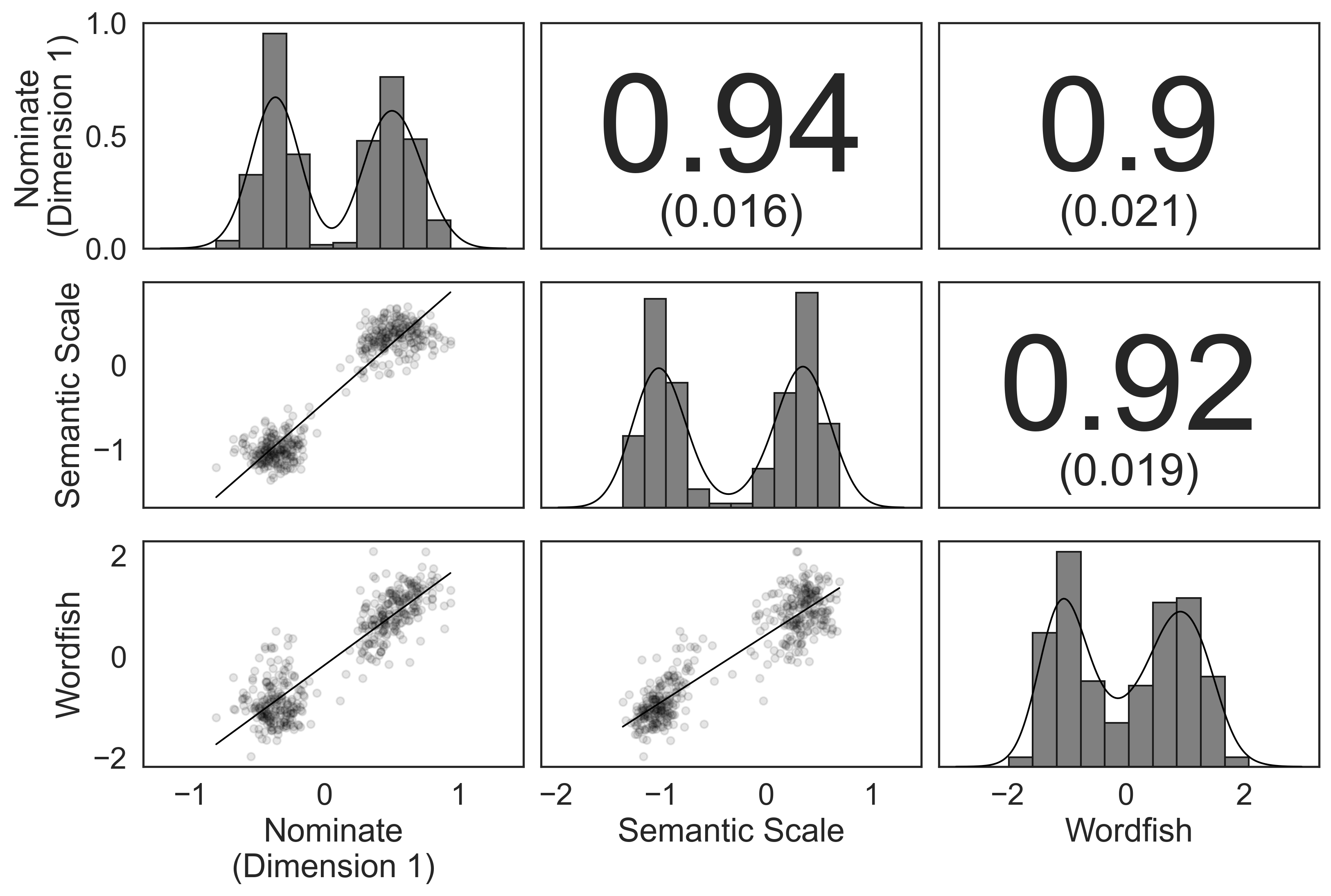}
    \caption{Scaling results using Twitter data only. Wordfish has reasonably high correlation with DW-NOMINATE, but Semantic Scaling offers a significant improvement and better separation between parties.}
    \label{fig:fish_twitter}
\end{figure}

While other researchers cited in section two of the paper have done more complete validations of other text-based scaling methods, I offer a brief comparison between Wordfish and Semantic Scaling here. Figure \ref{fig:fish_letters} compares scaling results when only using the congressional newsletters data and figure \ref{fig:fish_twitter} presents results using Twitter data only. Comparing the two data sources with different document types and language offers an informative contrast between semantic scaling and bag-of-words based scaling methods. Semantic scaling provides consistent results regardless of the document type. Conversely, the performance of Wordfish dramatically decreases when estimating ideology from newsletters and it fails to obtain meaningful separation between the parties. While Wordfish has reasonably high correlation with DW-NOMINATE on Twitter, it incorrectly identifies many Democrats as more conservative than their Republican colleagues and vice-versa. However, Semantic Scaling offers a significant improvement with perfect separation between the parties.

\clearpage
\section{Twitter Moderates}
One particularly interesting distinction between Semantic Scaling and Tweetscores is that Semantic Scaling is based on explicitly stated preferences while Tweetscores assumes the decision to follow a political figure is a latent expression of beliefs. Both approaches may be an advantage or disadvantage depending on the context. Ideological minorities may not feel comfortable expressing their opinions and thus Semantic Scaling may interpret this lack of partisan speech as evidence of moderate beliefs. Similarly, individuals may choose to follow political elites across the spectrum in order to stay informed and Tweetscores may be prone to label these individuals as moderates even though this is not necessarily true.

This makes differences in the moderates identified by each approach a particularly interesting point of comparison to identify strengths and weaknesses. To investigate this further I assign liberal, conservative, and moderate labels to each observation in my sample. To define moderates I locate the local minimum between the distributions and the area between that minimum and the peak of the modes. I consider moderates to be observations that occupy the 25\% of this area that is closest to the minimum. Observations to the left of this area are classified as liberals and observations to the right are classified as conservatives.

\begin{figure}[h]
    \centering
    \includegraphics[width = \textwidth]{Figures/confusion matrix.png}
    \caption{Confusion matrix between Semantic Scaling and Tweetscores.}
    \label{fig:twit_confusion}
\end{figure}

Figure \ref{fig:twit_confusion} uses these labels to derive a confusion matrix between Tweetscores and Semantic Scaling. From here we can estimate if a particular ideological group is over or underestimated by one of the methods. Assume the 507 observations that both Semantic Scaling and Tweetscores agree are moderate constitute ``true'' moderates and the remaining are ``false'' moderates. Assuming miss-classification as a moderate is random, what would be the expected count of conservatives as measured by Tweetscores/Semantic Scaling among the moderates as measured by Semantic Scaling/Tweetscores? For example, the counts of Tweetscore conservatives and liberals among Semantic Scaling moderates should be proportional to the total count of Tweetscore conservatives and liberals. If it is not, this suggests a potential bias in the estimates. 

In the context of Twitter, we might expect Semantic Scaling to disproportionately classify conservatives as moderates because they constitute an ideological minority that may be more hesitant to express their beliefs \citep{barbera2015birds, preoctiuc2017beyond}. Conversely, we might expect Tweetscores to disproportionately classify liberals as moderates because liberals are more likely to have heterophilous social networks and engage with others across the ideological divide \citep{boutyline2017social, barbera2015birds}. To determine if this is the case I estimate the expected count of Tweetscore conservatives among the Semantic Scale moderates and the number of Semantic Scale liberals among the Tweetscore moderates and test if the observed counts fall within a 95\% confidence interval of the expected counts.

\begin{figure}
    \centering
    \includegraphics[width=\textwidth]{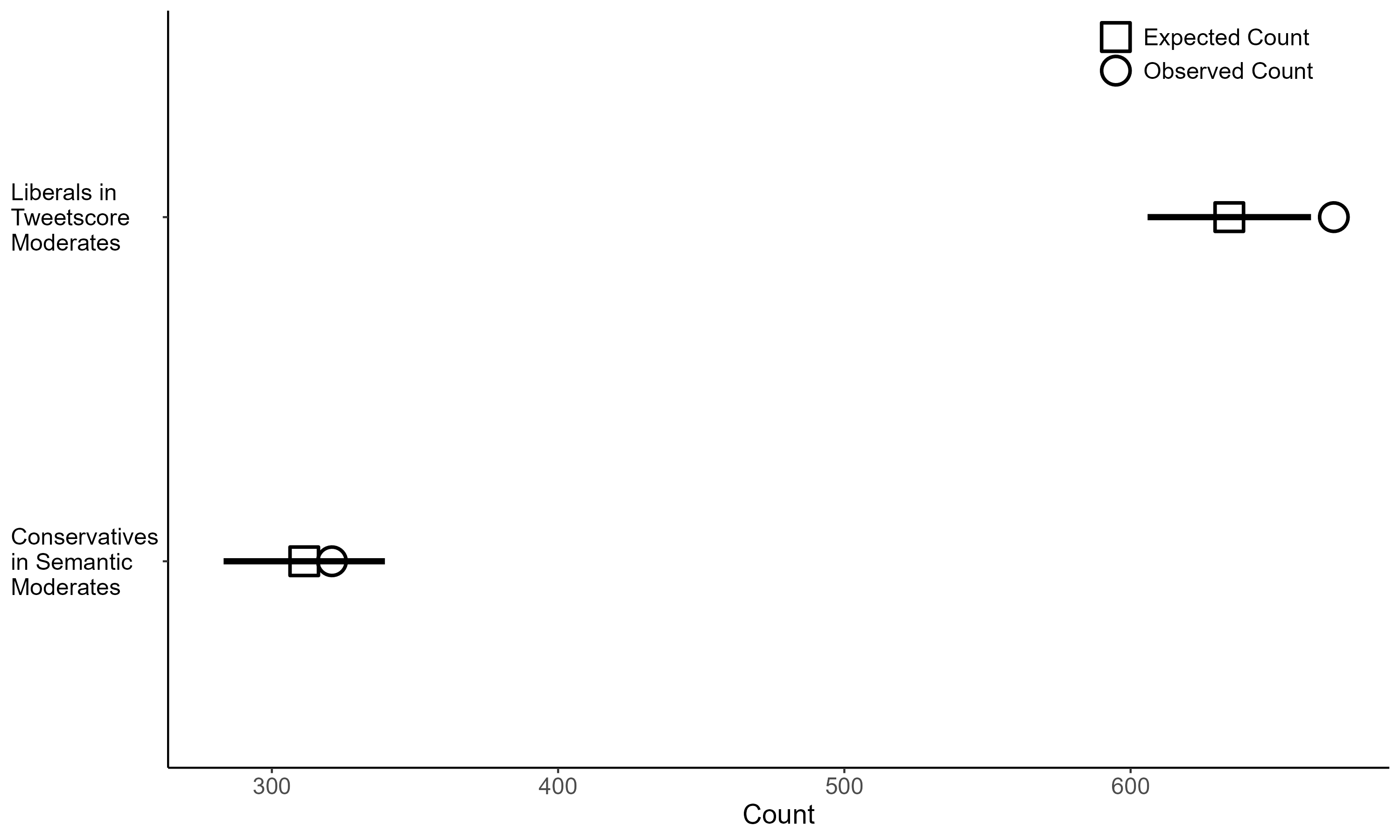}
    \caption{On Twitter, Tweetscores is more likely to estimate liberals as moderates than Semantic Scaling. There is no evidence that Semantic Scaling is under-counting conservatives in my sample, but this may be due to the fact that users must make a minimum number of posts to be included in my sample. My sample is thus biased towards including those that are comfortable sharing their opinion.}
    \label{fig:twit_moderates}
\end{figure}

The results of these hypothesis tests are shown in figure \ref{fig:twit_moderates}. I find that the observed count of Semantic Scale liberals among Tweetscore moderates is outside of the confidence interval, providing evidence that Tweetscores may indeed be under-counting liberals. However, I find no evidence that Semantic Scaling is underestimating the count of conservatives. It is very possible that this is due to sampling bias. Because observations must have a minimum number of posts in order to be included in my sample, my sample likely consists of users that are more comfortable sharing their opinions than the general public. 
\clearpage
\section{Entailment Classification Targets and Hypotheses}
\subsection{Hypotheses and Targets for Application 1: Twitter}
For items relating to affect, each tweet was classified for textual entailment using the following hypothesis template: ``The Author of this tweet (supports/opposes/does not express an opinion about) {target}" where target is the name of a person or group. Targets are listed below. Hypotheses were matched to targets based on keyword matching. For example, a tweet containing the keyword ``Biden" would be matched with the hypothesis: ``The Author of this tweet (supports/opposes/does not express an opinion about) Biden."
Items related to policy dimensions were classified using the following hypothesis template: ``The author of this tweet believes {belief statement}" where the belief statement is a simple phrase expressing a political opinion (e.g. ``gun control is good"). A list of affect targets and policy keyword-hypothesis pairs is below.

Affect Targets:
\begin{multicols}{2}
\small
\begin{itemize}
    \item Biden
    \item liberals
    \item leftists
    \item democrats
    \item Stewart
    \item Oliver
    \item Maddow
    \item Hayes
    \item O'Donnell
    \item Klein
    \item Krugman
    \item Thunberg
    \item Obama
    \item Clinton
    \item Sanders
    \item Pelosi
    \item Schiff
    \item Warren
    \item Omar
    \item Cortez
    \item Booker
    \item Schumer
    \item Kamala
    \item Trump
    \item Republicans
    \item Conservatives
    \item Shapiro
    \item Carlson
    \item Kirk
    \item D'Souza
    \item Ingraham
    \item Hannity
    \item Walsh
    \item Peterson
    \item Crowder
    \item Beck
    \item Malkin
    \item Rubio
    \item Paul
    \item Cruz
    \item Jordan
    \item McCarthy
    \item Scott
    \item Scalise
    \item Giuliani
    \item Ivanka
    \item Hawley
    \item McConnell
    \item Bush
    \item Reagan
\end{itemize}
\end{multicols}

\begin{table}[ht]
\small % Set font size to small
\centering
\begin{tabular}{>{\raggedright}p{4.5cm}p{8.5cm}} % Adjust column widths
\toprule
\textbf{Policy Keywords} & \textbf{Policy Hypotheses} \\
\midrule
flu & \makecell[tl]{covid is like the flu \\ covid is different than the flu \\ nothing about covid and the flu} \\
mask & \makecell[tl]{masks are good \\ masks are bad \\ masks neutral} \\
shutdown & \makecell[tl]{lockdowns are good \\ lockdowns are bad \\ lockdowns neutral} \\
vaccin & \makecell[tl]{vaccines are good \\ vaccines are bad \\ vaccine neutral} \\
social\_distan & \makecell[tl]{social distancing is good \\ social distancing is bad \\ social distancing neutral} \\
pandemic & \makecell[tl]{pandemic is dangerous \\ pandemic is not dangerous \\ pandemic neutral} \\
covid & \makecell[tl]{covid is dangerous \\ covid is not dangerous \\ covid neutral} \\
gun\_control & \makecell[tl]{gun control is good \\ gun control is bad \\ gun control neutral} \\
gun\_violence & \makecell[tl]{gun violence is a problem \\ gun violence is not a problem \\ gun violence neutral} \\
gun\_law & \makecell[tl]{gun laws are good \\ gun laws are bad \\ gun laws neutral} \\
second\_amendment & \makecell[tl]{the second amendment is good \\ the second amendment is bad \\ the second amendment neutral} \\
\bottomrule
\end{tabular}
\end{table}

\clearpage

\begin{table}[ht]
\small % Set font size to small
\centering
\begin{tabular}{>{\raggedright}p{4.5cm}p{8.5cm}} % Adjust column widths
\toprule
\textbf{Policy Keywords Cont.} & \textbf{Policy Hypotheses Cont.} \\
\midrule
black\_lives\_matter & \makecell[tl]{black lives matter is good \\ black lives matter is bad \\ black lives matter neutral} \\
george\_floyd & \makecell[tl]{floyd is good \\ floyd is bad \\ floyd neutral} \\
police & \makecell[tl]{police are good \\ police are bad \\ police are neutral} \\
white\_supremac & \makecell[tl]{trump, republicans, or conservatives are white supremacist \\ trump, republicans, or conservatives are not white supremacist \\ white supremacy neutral} \\
systemic\_racism & \makecell[tl]{systemic racism exists \\ systemic racism does not exist} \\
critical\_race\_theory & \makecell[tl]{critical race theory is good \\ critical race theory is bad \\ critical race theory neutral} \\
mail\_in & \makecell[tl]{mail in ballots are good \\ mail in ballots are bad \\ mail in ballots neutral} \\
voter\_id & \makecell[tl]{voter ids are good \\ voter ids are bad \\ voter ids neutral} \\
voter\_fraud & \makecell[tl]{voter fraud is real \\ voter fraud is not real} \\
stop\_the\_steal & \makecell[tl]{stop the steal good \\ stop the steal bad \\ stop the steal neutral} \\
election\_steal & \makecell[tl]{the election was stolen \\ the election was not stolen \\ trump tried to steal the election \\ the election was stolen from trump} \\
election\_fraud & \makecell[tl]{election fraud is real \\ election fraud is not real} \\
abortion & \makecell[tl]{abortion rights are good \\ abortion rights are bad \\ abortion is neutral} \\
planned\_parenthood & \makecell[tl]{planned parenthood is good \\ planned parenthood is bad \\ planned parenthood is neutral} \\
pro\_choice & \makecell[tl]{pro choice is good \\ pro choice is bad \\ pro choice neutral} \\
pro\_life & \makecell[tl]{pro life is good \\ pro life is bad \\ pro life neutral} \\
roe & \makecell[tl]{roe is good \\ roe is bad \\ roe neutral} \\
\bottomrule
\end{tabular}
\end{table}

\clearpage

\subsection{Hypotheses and Targets for Application 2: 117th Congress}
For application two I followed a similar classification procedure as I did with the Twitter application. A list of affect targets and policy keyword-hypotheses pairs are below.

\begin{multicols}{2}
\small
\begin{itemize}
    \item Trump
    \item Pence
    \item Biden
    \item Harris
    \item McConnell
    \item Schumer
    \item McCarthy
    \item Pelosi
    \item Liberals
    \item Democrats
    \item Conservatives
    \item Republicans
\end{itemize}
\end{multicols}

\begin{table}[ht]
\small
\centering
\begin{tabular}{>{\raggedright}p{5cm}p{10cm}}
\toprule
\textbf{Policy Keywords} & \textbf{Policy Hypotheses} \\
\midrule
abortion & \makecell[tl]{believes abortion is good \\ believes abortion is bad \\ believes abortion is neutral} \\
planned parenthood & \makecell[tl]{believes planned parenthood is good \\ believes planned parenthood is bad \\ believes planned parenthood is neutral} \\
pro-choice & \makecell[tl]{believes pro-choice is good \\ believes pro-choice is bad \\ believes pro-choice neutral} \\
pro-life & \makecell[tl]{believes pro-life is good \\ believes pro-life is bad \\ believes pro-life neutral} \\
gun & \makecell[tl]{believes guns are good \\ believes guns are bad \\ believes guns neutral} \\
second amendment & \makecell[tl]{believes the second amendment is good \\ believes the second amendment is bad \\ believes the second amendment neutral} \\
border & \makecell[tl]{believes there is a border crisis \\ believes there is not a border crisis \\ says nothing about the border crisis} \\
immigra & \makecell[tl]{supports immigration \\ opposes immigration \\ immigration neutral} \\
Inflation Reduction Act & \makecell[tl]{supports Inflation Reduction Act \\ opposes Inflation Reduction Act} \\
Infrastructure Investment and Jobs Act & \makecell[tl]{supports Infrastructure Investment and Jobs Act \\ opposes Infrastructure Investment and Jobs Act} \\
Build Back Better Act & \makecell[tl]{supports Build Back Better Act \\ opposes Build Back Better Act} \\
American Rescue Plan Act & \makecell[tl]{supports American Rescue Plan Act \\ opposes American Rescue Plan Act} \\
CARES Act & \makecell[tl]{supports CARES Act \\ opposes CARES Act} \\
Honoring our PACT Act & \makecell[tl]{supports Honoring our PACT Act \\ opposes Honoring our PACT Act} \\
Respect for Marriage Act & \makecell[tl]{supports Respect for Marriage Act \\ opposes Respect for Marriage Act} \\
For the People Act & \makecell[tl]{supports For the People Act \\ opposes For the People Act} \\
America COMPETES Act & \makecell[tl]{supports America COMPETES Act \\ opposes America COMPETES Act} \\
Voting Rights Advancement Act & \makecell[tl]{supports Voting Rights Advancement Act \\ opposes Voting Rights Advancement Act} \\
\bottomrule
\end{tabular}
\end{table}

\clearpage

\section{MCMC Convergence and Diagnostics}
Each model ran four chains. To ensure convergence across chains I ran each model for a pilot run of 4,000 iterations and then used the Raftery diagnostic to estimate the maximum iterations needed for a 0.005 margin of error with 95\% confidence. I found that 40,000 iterations for the Twitter models and 100,000 iterations for the congressional models were more than sufficient. Across all models and parameters, no parameter had a bulk effective sample size lower than the 400 recommended by \cite{vehtari2021rank} and no parameter had an $\hat{R}$ higher than the recommended 1.01. There were no divergent chains post warmup in any of the models and no iterations saturated the maximum tree depth of 10. Because the number of parameters are too numerous to present a full set of trace plots, I include a sample from each parameter for all models in the sections below.
\subsection{Twitter Affect Trace Plots}
\includegraphics[width=\textwidth]{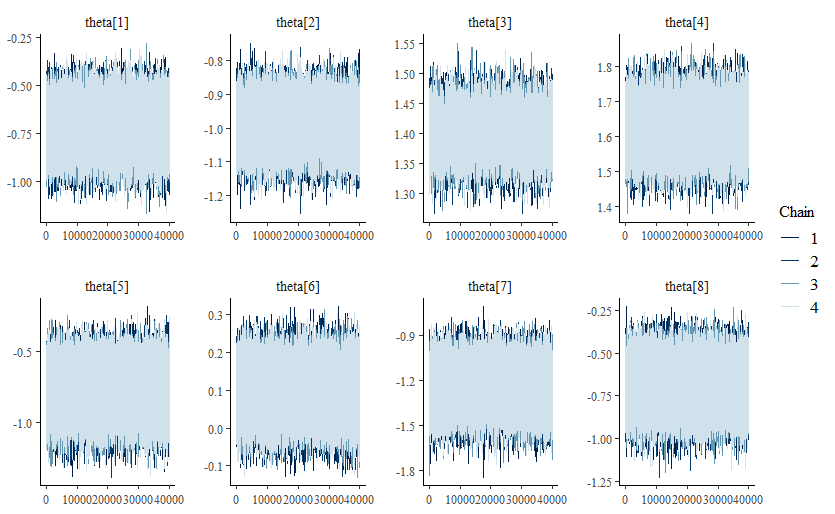}
\includegraphics[width=\textwidth]{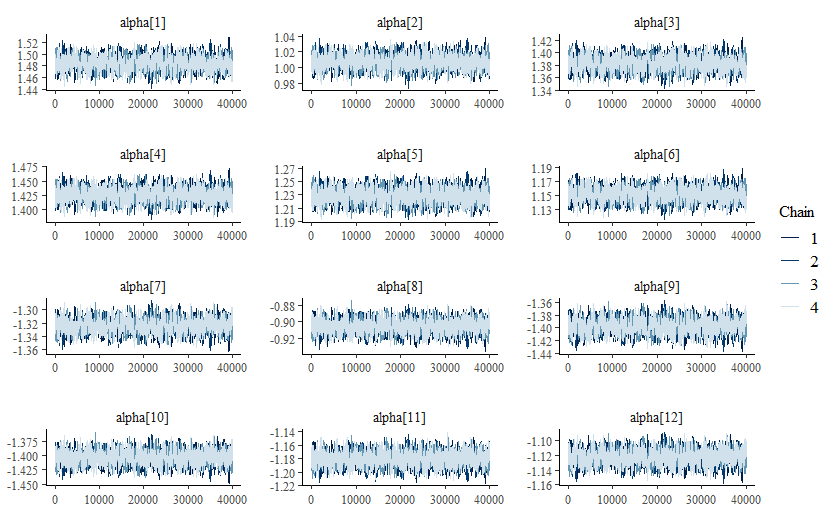}
\includegraphics[width=\textwidth]{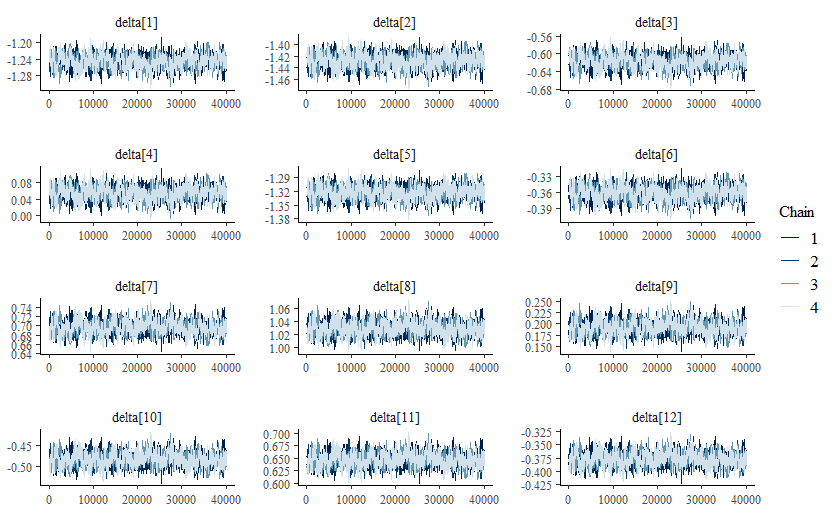}

\subsection{Twitter Policy Trace Plots}
\includegraphics[width=\textwidth]{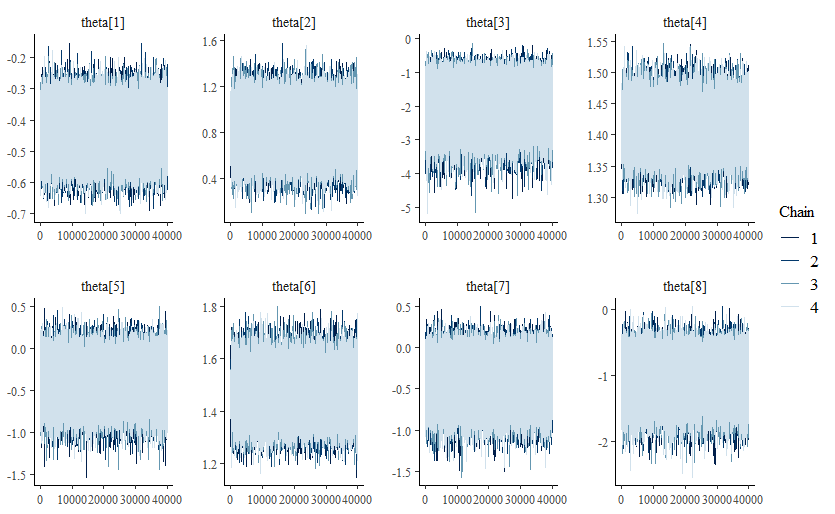}
\includegraphics[width=\textwidth]{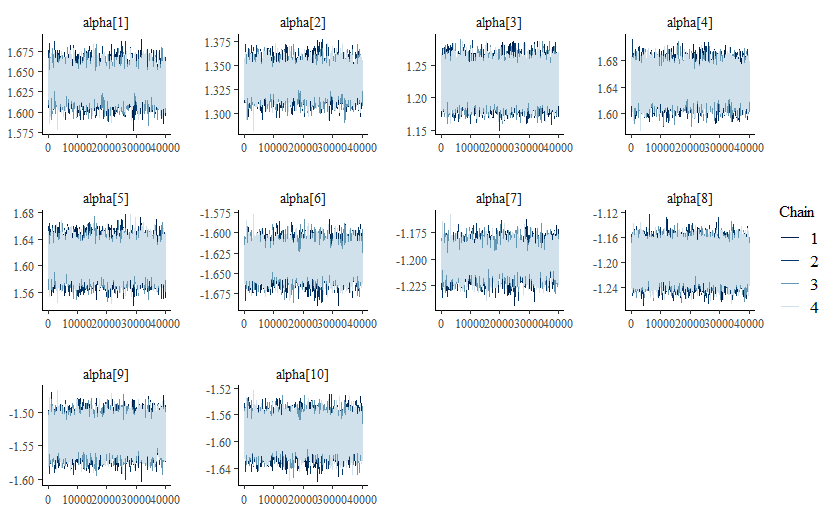}
\includegraphics[width=\textwidth]{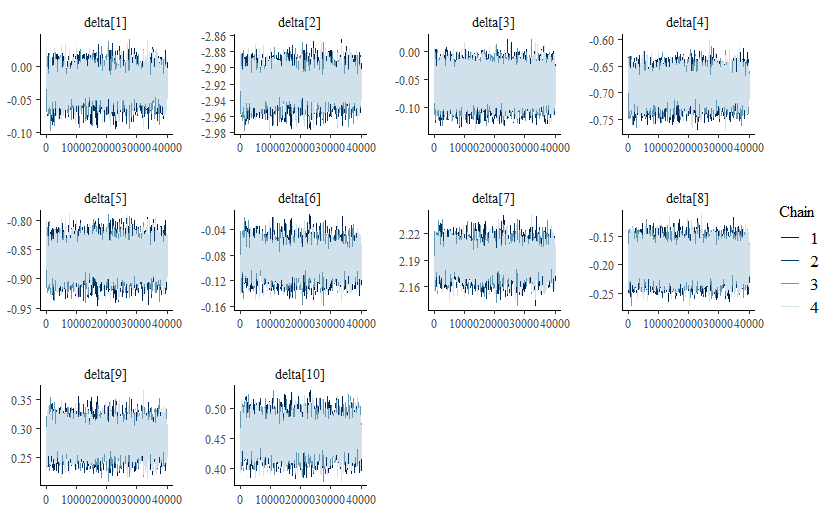}

\subsection{Twitter Joint Trace Plots}
\includegraphics[width=\textwidth]{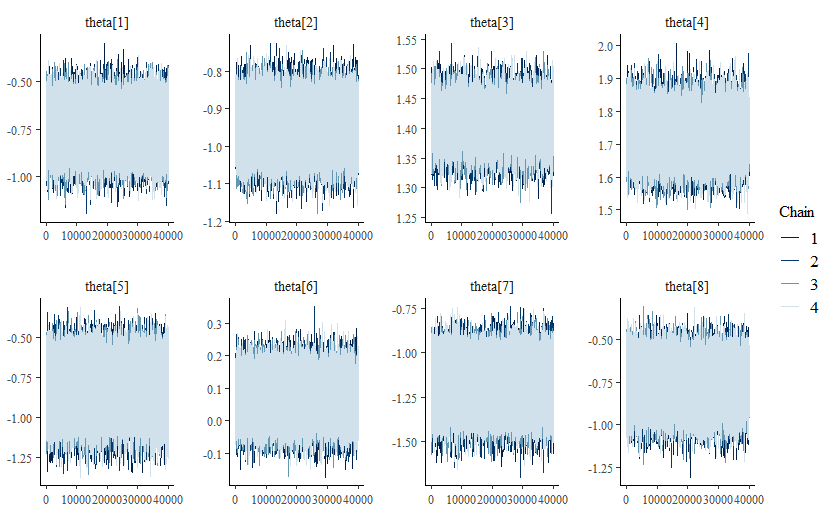}
\includegraphics[width=\textwidth]{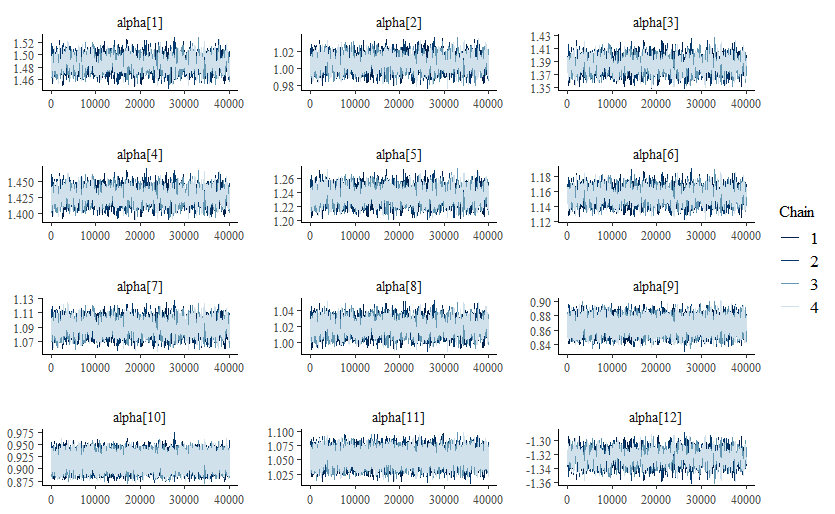}
\includegraphics[width=\textwidth]{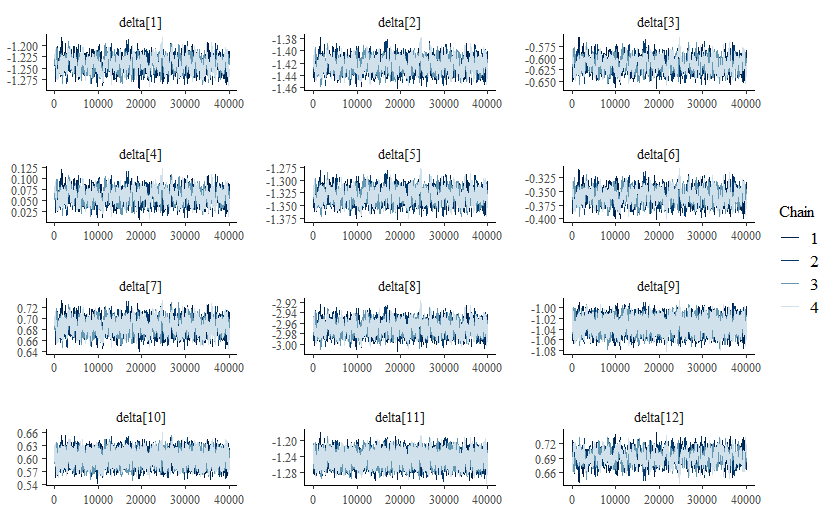}

\subsection{Congress Affect Trace Plots}
\includegraphics[width=\textwidth]{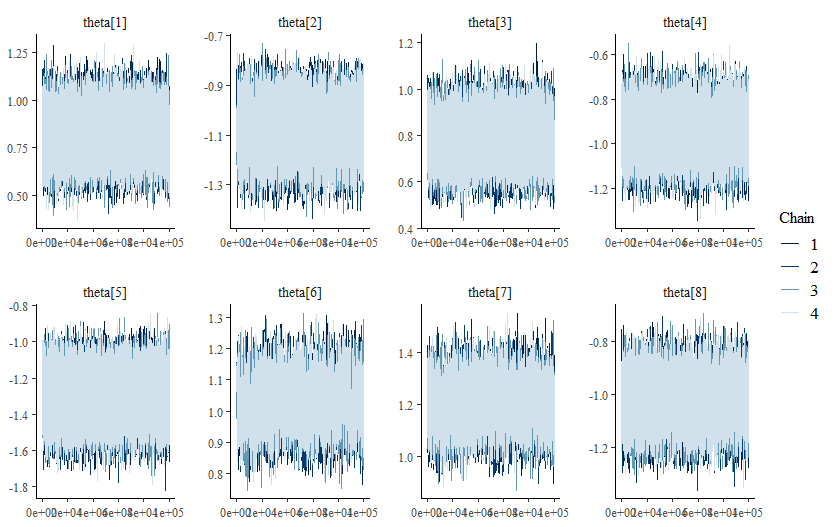}
\includegraphics[width=\textwidth]{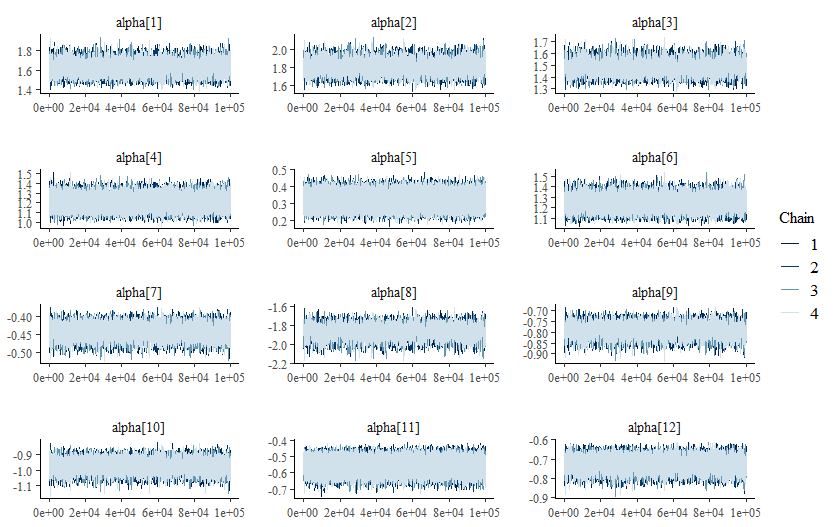}
\includegraphics[width=\textwidth]{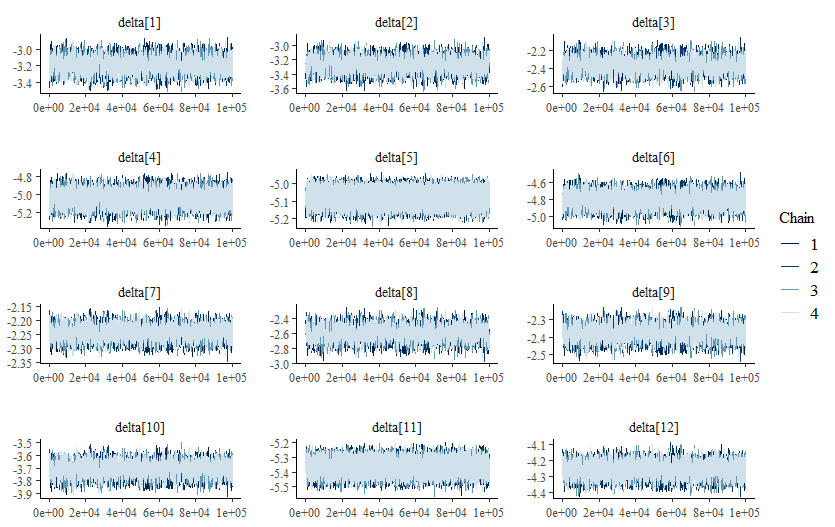}
\subsection{Congress Policy Trace Plots}
\includegraphics[width=\textwidth]{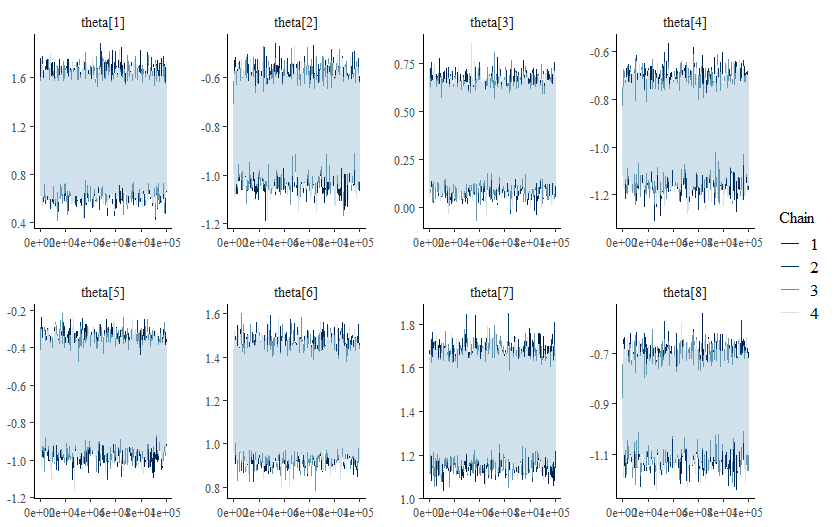}
\includegraphics[width=\textwidth]{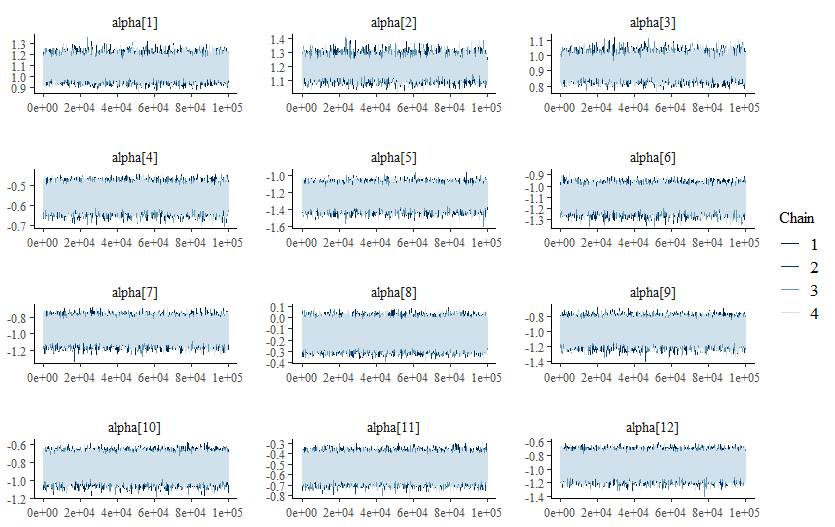}
\includegraphics[width=\textwidth]{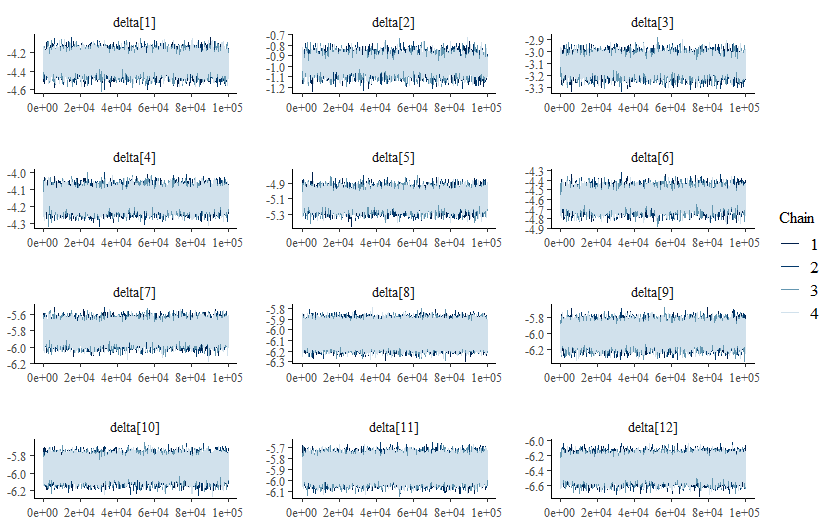}
\subsection{Congress Joint Trace Plots}
\includegraphics[width=\textwidth]{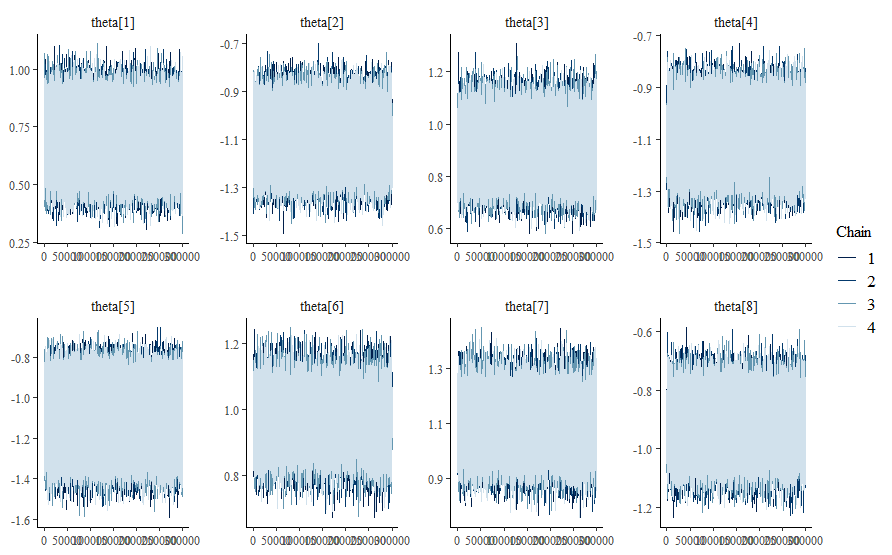}
\includegraphics[width=\textwidth]{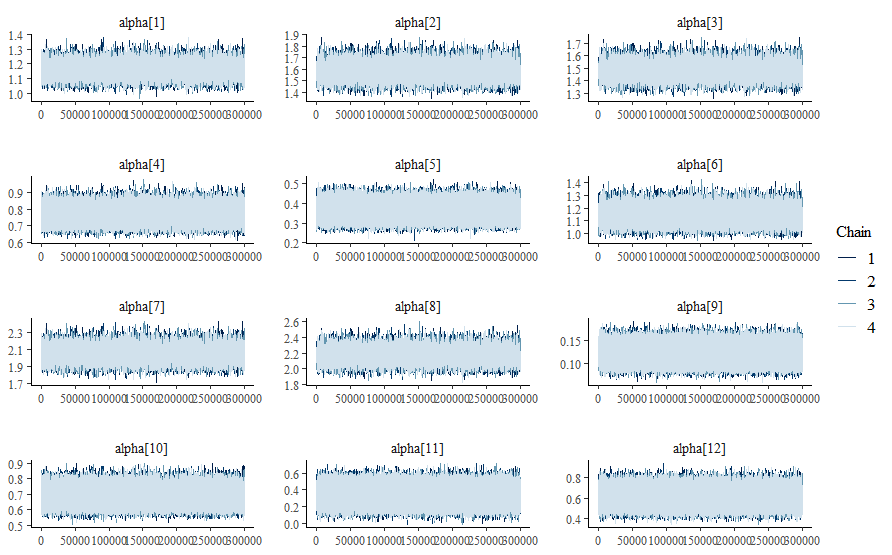}
\includegraphics[width=\textwidth]{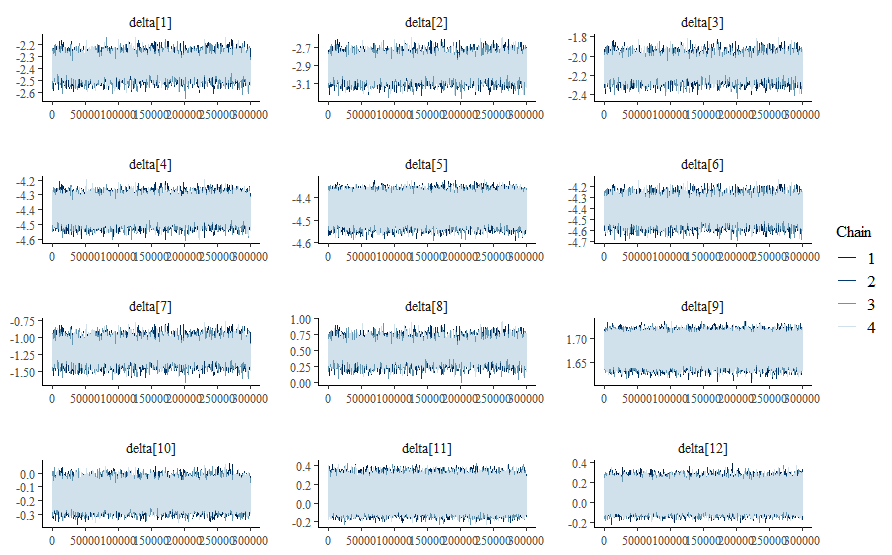}